\pdfoutput=1
\PassOptionsToPackage{table}{xcolor}
\documentclass{article}
\usepackage{iclr2027_conference,times}
\usepackage[
  letterpaper,
  left=1.15in,
  right=1.15in,
  top=0.78in,
  bottom=0.78in,
  headheight=12pt,
  headsep=0.18in,
  footskip=0.32in
]{geometry}
\iclrfinalcopy 
\usepackage[T1]{fontenc}
\usepackage[utf8]{inputenc}
\usepackage{amsmath,amsfonts,bm}

\def\eqref#1{equation~\ref{#1}}

\def\1{\bm{1}}

\DeclareMathAlphabet{\mathsfit}{\encodingdefault}{\sfdefault}{m}{sl}
\SetMathAlphabet{\mathsfit}{bold}{\encodingdefault}{\sfdefault}{bx}{n}

\usepackage{hyperref}
\usepackage{url}
\urldef{\projecturl}\url{https://driveanchor-meituan.github.io/}

\usepackage{booktabs}
\usepackage{amsfonts}
\usepackage{amsmath}
\usepackage{amssymb}
\usepackage{graphicx}
\usepackage{xcolor}
\usepackage{float}
\usepackage{algorithm}
\usepackage{algorithmic}
\usepackage{caption}
\usepackage{subcaption}
\usepackage{multirow}
\usepackage{enumitem}
\usepackage{microtype}
\usepackage{placeins}
\usepackage{flafter}
\usepackage{needspace}
\usepackage{eso-pic}
\AddToShipoutPictureFG{%
  \AtPageLowerLeft{\makebox[\paperwidth][c]{\raisebox{0.28in}{\small\thepage}}}}
\renewcommand{\eqref}[1]{\textup{(\ref{#1})}}
\makeatletter
\g@addto@macro\normalsize{%
  \setlength{\abovedisplayskip}{6pt plus 2pt minus 3pt}%
  \setlength{\belowdisplayskip}{6pt plus 2pt minus 3pt}%
  \setlength{\abovedisplayshortskip}{2pt plus 2pt}%
  \setlength{\belowdisplayshortskip}{4pt plus 2pt minus 2pt}%
}
\makeatother
\definecolor{mtyellow}{HTML}{FCD000}
\definecolor{TabHead}{HTML}{F5F6F8}
\definecolor{TabGroup}{HTML}{FAFAFB}
\definecolor{TabOurs}{HTML}{F2F7FD}
\usepackage{etoolbox}
\AtBeginEnvironment{table}{\setlength{\aboverulesep}{0pt}\setlength{\belowrulesep}{0pt}}
\definecolor{mtamber}{HTML}{B36B00}
\definecolor{mtdark}{HTML}{1A1A1A}
\hypersetup{colorlinks=true,linkcolor=mtamber,citecolor=mtamber,urlcolor=mtamber,filecolor=mtamber,
  pdftitle={DriveAnchor: Progressive Anchor-based Flow Learning for Autonomous Driving Planning},
  pdfauthor={Limin Yan; Haoyun Tang; Yutao Qiu; Hongqing Liu; Zhandong Mei; Haoyu Xu}}
\renewcommand{\headrulewidth}{0pt}
\makeatletter
\def\@maketitle{\vbox{\hsize\textwidth
  \noindent\makebox[\textwidth][s]{%
    \raisebox{3pt}{\includegraphics[height=20pt]{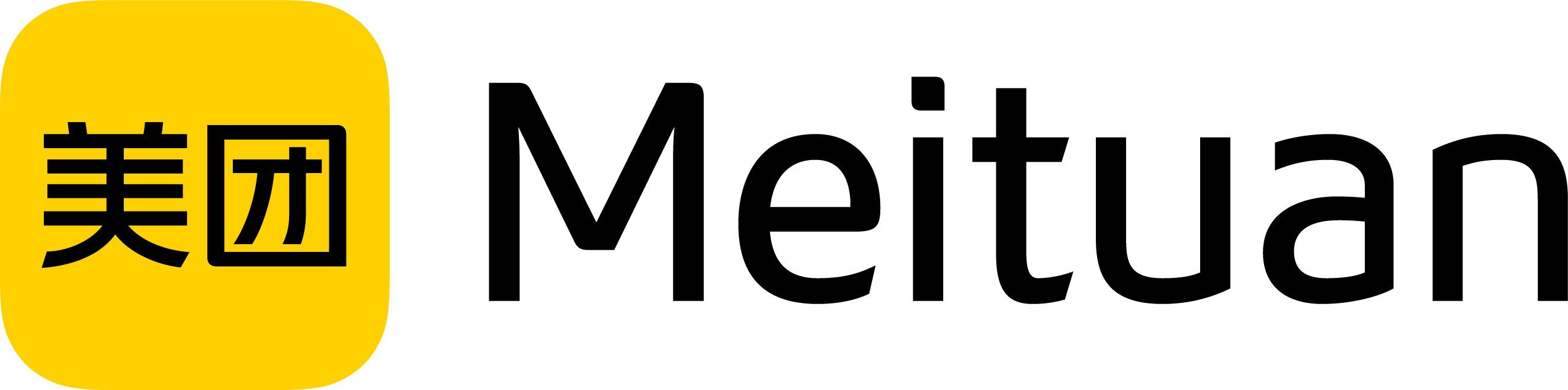}}\hfill
    \includegraphics[height=30pt]{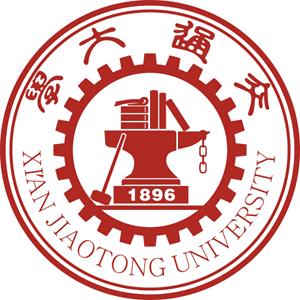}\hspace{9pt}%
    \includegraphics[height=30pt]{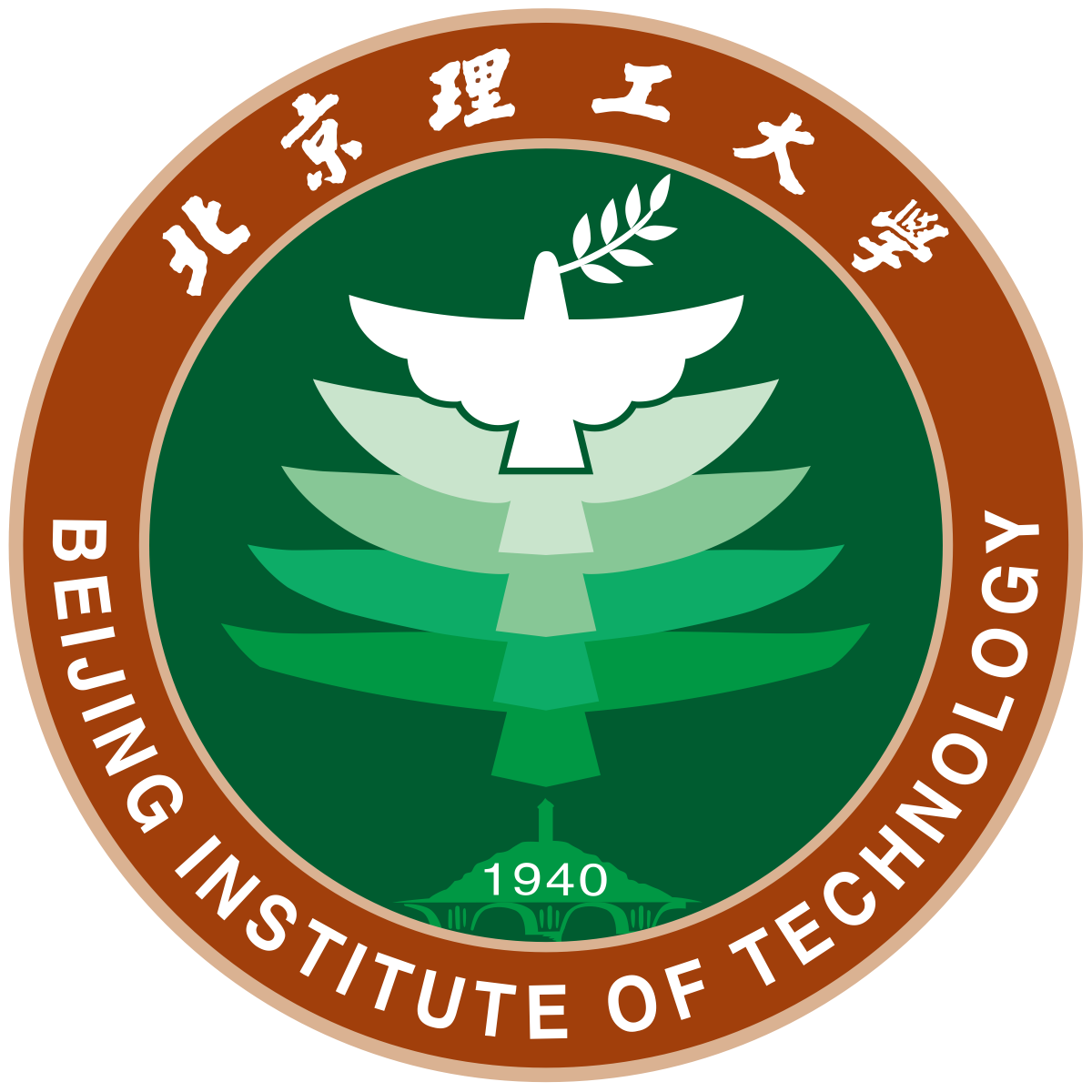}}\par
  \vskip 6pt
  {\color{mtyellow}\hrule height 1.5pt}
  \vskip 0.28in
  \centering
  {\LARGE\bfseries\color{mtdark}\@title\par}
  \vskip 0.16in
  {\color{mtdark}\hrule height 0.6pt}
  \vskip 0.14in
  {\@author\par}
  \vskip 0.12in}}
\makeatother

\title{DriveAnchor: Progressive Anchor-based Flow Learning\\for Autonomous Driving Planning}
\author{%
{\bfseries
Limin Yan$^{1*\dagger}$,
Haoyun Tang$^{1,2*}$,
Yutao Qiu$^{1,3}$,
Hongqing Liu$^{1}$,
Zhandong Mei$^{2}$,
Haoyu Xu$^{1\S}$}\\[9pt]
$^{1}$Meituan\qquad
$^{2}$Xi'an Jiaotong University\qquad
$^{3}$Beijing Institute of Technology\\[7pt]
{\small\itshape Corresponding authors:}\enspace
{\small\ttfamily yanlimin03@meituan.com\quad xuhaoyu06@meituan.com}\\[4pt]
{\small\itshape Project page:}\enspace \projecturl%
}

\begin{document}
\maketitle
{\renewcommand{\thefootnote}{}\footnotetext{$^{*}$Equal contribution.\quad $^{\dagger}$Lead corresponding author.\quad $^{\S}$Co-corresponding author.}}
\setcounter{footnote}{0}
\raggedbottom
\brokenpenalty=10000

\begin{abstract}

Autonomous-driving planners must generate route- and maneuver-compliant trajectories under a limited decoder-call budget while adapting offline to black-box static-collision feedback. We propose \textbf{DriveAnchor}, a three-stage framework built around a shared vocabulary of recorded trajectories and a continuous trajectory interface. The vocabulary supports flow-matching (FM) residual learning, Energy Field (EF) corridor-conditioned initialization, and anchor-guided zeroth-order reward fine-tuning. The reward update operates in trajectory space without policy likelihoods or reward gradients, enabling offline adaptation of a deterministic decoder. On sampled nuPlan scenarios, EF guidance and reward fine-tuning each improve the mean closed-loop score. On the same cases, the official pretrained Diffusion Planner scores higher with our reward update than with supervised fine-tuning. We further integrate the trajectory generator into a real vehicle's planning pipeline and demonstrate its operation on urban roads. On 10,000 held-out internal scenarios, under the original non-EF prior and a fixed two-call FM budget, reward fine-tuning reduces short- and long-horizon static candidate collision rates from 27.22\% and 56.54\% to 2.94\% and 7.15\%, respectively, and improves 8-s best-of-set minADE from 2.27 to 1.34\,m. In a separate junction diagnostic on checkpoints before Stage~3, one and two EF calls raise corridor-guidance satisfaction from 22.10\% to 74.13\% and 87.92\%, respectively. Additional ablations support the vocabulary size and the anchor-guided sampling configuration.
On NVIDIA Drive Orin, two EF calls and two FM calls take 2.06\,ms per scene on average. Code and deployment videos are available on the project page.
\end{abstract}

\section{Introduction}
\label{sec:intro}

\looseness=-1 A driving planner must generate trajectories that follow route and maneuver requests, select one for execution, and operate within a limited computation budget. A logged ego trajectory records one realized behavior, while the same scene may admit several plausible alternatives. Even a planner with a high mean score on a fixed simulation benchmark can fail in a new situation on the road. After deployment, developers need to address such failures promptly, often with only a few logs, since collecting corrective demonstrations takes time. These requirements call for a planning framework that connects diverse trajectory generation, explicit control, feedback-driven adaptation, and deployment.

These requirements meet at the trajectory interface. A control request changes where the decoder should start, while collision feedback indicates how its outputs should improve. For deterministic decoders, learning from this feedback is challenging: black-box collision checks may be non-differentiable, and the decoder does not directly provide the policy likelihoods used by stochastic reward-fine-tuning methods~\citep{black2024ddpo,zhang2025reinflow,xue2025dancegrpo,liu2025flowgrpo}. Exploration must also remain possible when the current model generates similar candidates.

\textbf{DriveAnchor} integrates scene encoding, controllable generation, reward adaptation, and candidate selection through a shared vocabulary of recorded trajectories. During FM training, locally perturbed anchors form trajectory starting states, so local anchor--demonstration pairs teach the decoder corrections across this structured distribution.

The three stages use this representation to learn FM residual corrections, construct EF corridor-conditioned starts, and explore reward improvements around decoded outputs. At deployment, EF and FM compose through a small number of forward calls, followed by candidate selection.

Our contributions are threefold:

\noindent\textbf{An integrated framework for controllable planning and targeted adaptation.} A common trajectory interface connects supervised generation, corridor control, and black-box static-collision reward learning, and allows the corridor request or the decoder alone to be adapted.

\noindent\textbf{A shared anchor prior for generation and reward exploration.} Anchors serve as noisy FM starting states and provide reward-query directions independently of decoder-output diversity. We derive a likelihood-free trajectory-space update and characterize its expectation through a local, directionally smoothed surrogate objective in trajectory space.

\noindent\textbf{System evaluation and controlled analysis.} We evaluate complete planners on matched nuPlan scenarios, demonstrate urban-road deployment with Orin runtime measurements, and use controlled internal experiments to assess Stage~3 and EF.

\section{Related Work}
\label{sec:related}

\paragraph{Generative planning and structured priors.}
Diffusion and flow-matching models~\citep{lipman2023flow,liu2023flow,albergo2023stochastic,chen2024riemannian} have been applied to trajectory generation and planning~\citep{janner2022diffuser,zheng2025diffusionplanner,tan2025flowplanner}. End-to-end planners combine scene representations with trajectory prediction~\citep{hu2023uniAD,jiang2023vad,cheng2024plantf}, while anchor-based methods represent possible futures through candidate trajectories, goals, or intentions~\citep{chai2020multipath,zhao2021tnt,shi2022mtr}. Goal-conditioned Transformers unify prediction and planning with feasibility and interaction terms~\citep{selzer2026plantransformer}, and uncertainty-aware modules adapt prediction and ego planning to challenging agents in hybrid learning/rule-based planners~\citep{chen2025uncertainty}. DiffusionDrive combines multimodal anchors with truncated denoising~\citep{liao2025diffusiondrive}, and DiffusionDriveV2 uses intra- and inter-anchor GRPO to improve generated trajectories~\citep{zou2025diffusiondrivev2}. SparseDrive selects among parallel planning candidates with a collision-aware rescoring module~\citep{sun2025sparsedrive}, and BridgeDrive formulates the refinement from a coarse anchor to a plan as a diffusion bridge~\citep{liu2025bridgedrive}. DriveAnchor trains FM from locally perturbed trajectory anchors and reuses the vocabulary to guide corridor-based initialization and define exploration directions for trajectory-space reward updates.

\paragraph{Controllable generation.}
Classifier-free guidance combines conditional and unconditional predictions to steer generation~\citep{ho2021classifierfree}, while cost-based guidance uses objective gradients during sampling~\citep{zhong2023ctg,zhong2023ctgpp}. Int2Planner conditions multimodal planning on route intention points and has been deployed on real vehicles~\citep{chen2025int2planner}. DriveAnchor instead learns a correction conditioned on the corridor geometry and current trajectory, and draws multimodality from the trajectory vocabulary rather than from intention points. EF is trained by regression and guides initialization through a forward pass at deployment.

\paragraph{RL fine-tuning of generative models.}
Reward-aligned fine-tuning includes RLHF~\citep{ouyang2022instructgpt} and GRPO~\citep{shao2024deepseekmath}, which estimates advantages within a sampled group and optimizes a clipped likelihood-ratio objective. DDPO~\citep{black2024ddpo}, ReinFlow~\citep{zhang2025reinflow}, and DanceGRPO~\citep{xue2025dancegrpo} optimize stochastic generative policies without differentiating the reward function. Flow-GRPO~\citep{liu2025flowgrpo} introduces stochastic flow transitions through an ODE-to-SDE conversion. Applying these approaches requires a probability model that our deterministic FM decoder does not directly provide. MeZO~\citep{malladi2023mezo} instead studies zeroth-order fine-tuning in parameter space, ZPG estimates policy gradients from preference queries without reward inference~\citep{zhang2025zerothorder}, and Guided ES~\citep{maheswaranathan2019guided} shapes the search distribution using surrogate-gradient directions. RL has also been applied directly to driving planners: CarPlanner trains a consistent autoregressive policy with a generation-selection scheme at nuPlan scale~\citep{zhang2025carplanner}, and NeuralParker learns a continuous-trajectory parking policy validated on a real vehicle~\citep{wang2026neuralparker}. ReCogDrive fine-tunes a diffusion planner with group-relative advantages on denoising-transition likelihoods~\citep{li2025recogdrive}, and Raw2Drive trains actor--critic driving policies through aligned world-model rollouts~\citep{yang2025raw2drive}. DriveAnchor instead constructs exploration directions from nearby vocabulary trajectories and propagates the resulting output-space signal to the model parameters through the decoder Jacobian, without policy likelihoods or environment rollouts.

\section{Method}
\label{sec:method}

\begin{figure}[htbp]
    \centering
    \includegraphics[width=\linewidth]{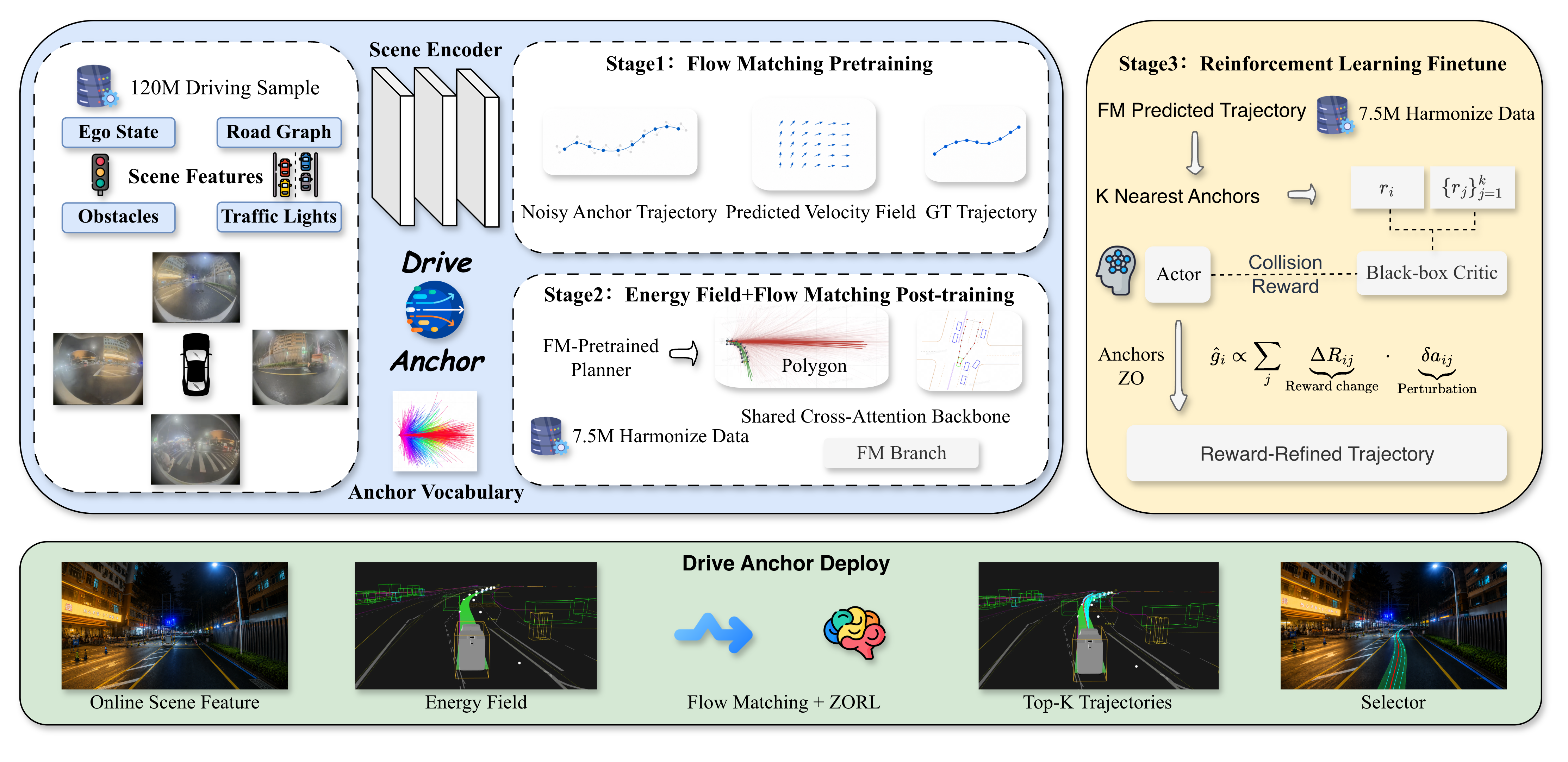}
    \caption{DriveAnchor training and deployment for the internal experiments. \textbf{Stage~1} (120M samples) learns residual corrections from perturbed local anchor--GT pairs. \textbf{Stage~2} (7.5M samples) buckets the data by dedicated scenario type, continues training FM on the bucketed data, and trains EF per scenario; the two are trained jointly with separate supervision, and EF predictions are not used as FM training inputs. \textbf{Stage~3} (2.5M samples) perturbs FM outputs, queries their rewards, and fine-tunes FM with a stop-gradient ZO signal. At inference, EF supplies initial trajectories to the reward-fine-tuned FM decoder, followed by residual screening and downstream selection; reward queries and ZO updates are performed offline.}
    \label{fig:overview}
\end{figure}

Given a scene and a corridor request, DriveAnchor generates candidates by repeatedly correcting continuous trajectories. A fixed vocabulary provides reference trajectories for training and reward exploration; EF and FM share the trajectory representation passed between modules. At deployment, the scene is encoded once, EF adjusts the initialization, and FM refines it for downstream selection. Figure~\ref{fig:overview} shows the internal pipeline. All ZO reward queries and parameter updates occur offline. We use ``FM decoder'' for the learned module and ``residual correction'' for its inference operation.

\Needspace{10\baselineskip}
\subsection{Stage~1: FM Pretraining}
\label{sec:anchor_construction}

A shared scene-conditioned FM decoder is trained on local anchor--GT pairs induced by a vocabulary of $M{=}2{,}398$ anchors. The vocabulary is built by iterative farthest-point sampling~\citep{qi2017pointnetpp} over a historical corpus $\mathcal{H}$ of 100M+ frames:
\begin{equation}
    \mathcal{A} = \text{FPS}\!\left(\{a^*_k\}_{k=1}^{|\mathcal{H}|},\; M\right), \quad a^*_k \in \mathbb{R}^{T \times 2}
    \label{eq:fps}
\end{equation}
Trajectories are flattened to $2T$-dimensional vectors for distances and inner products. The corpus is temporally disjoint from training and evaluation data. For each GT trajectory $x_1$, we select its $K_{\rm GT}$ nearest vocabulary anchors as pairing partners; the local GT pairing count $K_{\rm GT}$ is the number of nearest anchors paired with each GT trajectory, fixed to $K_{\rm GT}=50$ throughout. Stage~1 constructs training pairs only from these anchors. We first perturb the selected anchor, $\tilde{x}_0=a+\varepsilon_{\rm in}$, and then sample the linear path:
\begin{equation}
    x_t = (1-t)\,\tilde{x}_0 + t\,x_1, \quad t \sim \text{Uniform}(0,1)
    \label{eq:training_interp}
\end{equation}
The perturbed anchor defines the path's starting state; scene features $s$ provide conditioning.

The noise $\varepsilon_{\rm in}$ used in FM training is sampled as follows. In Stages~1 and~2, noise is drawn uniformly from a ball, and a perturbed trajectory is accepted only if its nearest vocabulary anchor remains unchanged, so the perturbed start still belongs to the same anchor. Appendix~\ref{app:training_curves} discusses the sampling procedure further.

Let $F_\theta$ denote the FM decoder: it takes the current trajectory $x_t\in\mathbb{R}^{T\times2}$, the time $t\in[0,1]$, and the scene features $s$, and outputs a predicted remaining displacement $F_\theta(x_t,t,s)\in\mathbb{R}^{T\times2}$ toward the target trajectory $x_1$. The time-conditioned decoder learns this displacement using the regression term retained in all stages:
\begin{equation}
 \mathcal L_{\mathrm{res}}(\theta)
 =\mathbb E\bigl[\ell_{\mathrm{SL1}}(F_\theta(x_t,t,s),x_1-x_t)\bigr].
 \label{eq:residual_loss}
\end{equation}
Here $\ell_{\mathrm{SL1}}$ denotes the coordinate-mean SmoothL1 loss. For the linear-path velocity target $v^\star=x_1-\tilde x_0$, the residual satisfies
\begin{equation}
    x_1-x_t=(1-t)v^\star.
    \label{eq:residual_velocity_relation}
\end{equation}
At inference, each FM call updates the trajectory:
\begin{equation}
    T_\theta(x,s) \triangleq x+F_\theta(x,0,s).
    \label{eq:reset_map}
\end{equation}
Starting from a raw anchor or an EF output, we repeatedly apply $T_\theta$, using the previous output as input and resetting $t=0$ at each call. This performs iterative residual correction. Section~\ref{sec:guidance_calls} evaluates the effect of repeated calls.

\Needspace{4\baselineskip}
\subsection{Stage~2: EF Post-training}
\label{sec:energy_field}

EF predicts an initialization correction $\Delta a_{\mathrm{EF}}\in\mathbb R^{T\times2}$ from the current trajectory, corridor request, and static-geometry features of the shared scene encoder. The corridor polygon and designated exit specify the maneuver. EF outputs a correction vector rather than a scalar energy.

\paragraph{Control-request construction.}
The corridor polygon and designated exit are constructed from the navigation route, map geometry, and current ego state, independently of the recorded ground-truth trajectory; the ground truth need not satisfy the corridor request.
The six corridor presets and their geometric construction are detailed in Appendix~\ref{app:ef_polygons} (Figure~\ref{fig:app_ef_polygons}).

One EF call followed by one FM call gives:
\begin{equation}
    \tilde{a}_{\text{EF}} = a_{\text{anchor}} + \Delta a_{\text{EF}}, \qquad
    a_{\text{final}} = T_\theta(\tilde{a}_{\text{EF}},\, s).
    \label{eq:ef_step1}
\end{equation}
\looseness=-1 EF passes its continuous output directly to FM, without vocabulary rematching. Further FM calls follow Eq.~\eqref{eq:reset_map}.

\paragraph{Training and composition.}
Stage~2 first buckets the training data by dedicated scenario type, then continues training FM on the bucketed data and trains EF specifically for each scenario. The two are trained jointly with separate targets: corridor corrections for EF and anchor--demonstration residuals for FM. Both use a frozen scene encoder and a shared trainable trajectory projector. FM training does not use EF predictions as inputs. At inference, EF can refine its output repeatedly before passing it to FM.

Stage~3 decodes raw anchors during training. Composing EF with the trained FM decoder therefore requires generalization to EF-adjusted inputs. We test whether the EF+FM composition is effective through candidate quality after EF initialization and final-output guidance satisfaction.

\paragraph{Training objective.}
Let $\operatorname{touch}_{\mathcal P}(a)$ indicate that the trajectory enters the polygon or starts inside it, with an additional permitted-entry-side check for lane changes. Let $\operatorname{inside}_{\rm req}(a)$ indicate occupancy at a specified time when that optional check is enabled, $n_{\rm exit}(a)$ count exits, and $e_{\rm exit}(a)$ identify the edge used for a single exit. The time check is part of label construction, not an extra EF input, and is always true when disabled. The geometric label is
\begin{equation}
\begin{aligned}
 \operatorname{good}(a)
 &=\operatorname{touch}_{\mathcal P}(a)\wedge\operatorname{inside}_{\rm req}(a)\\
 &\quad\wedge\bigl[n_{\rm exit}(a)=0
 \ \vee\ (n_{\rm exit}(a)=1\wedge e_{\rm exit}(a)=\mathcal E_k)\bigr].
\end{aligned}
\label{eq:ef_label}
\end{equation}
A trajectory can satisfy the request by remaining inside the polygon after entry; if it leaves, it must use the designated exit. We evaluate the geometric criterion separately for unperturbed vocabulary entries and noisy inputs. The former define the reference pool $\mathcal A_{\rm good}=\{b\in\mathcal A:\operatorname{good}(b)\}$, while the latter determine which supervision branch to use. For a nonempty reference pool,
\begin{equation}
 \resizebox{0.88\linewidth}{!}{$\displaystyle \delta^*(a_{\rm noisy})=\begin{cases} \mathbf 0,&\operatorname{good}(a_{\rm noisy}),\\ b^*(a_{\rm noisy})-a_{\rm noisy},&\text{otherwise}, \end{cases}\quad b^*(a_{\rm noisy})\in\arg\min_{b\in\mathcal A_{\rm good}}\frac{1}{T}\sum_{t=1}^{T}\|b_t-a_{{\rm noisy},t}\|_2.$}
 \label{eq:ef_target}
\end{equation}
EF uses mean pointwise Euclidean distance to retrieve references and SmoothL1 with kinematic losses to learn their displacements.

\Needspace{4\baselineskip}
\subsection{Stage~3: RL Fine-tuning}
\label{sec:stage3}

Stage~3 samples starting anchors uniformly from the full vocabulary and adapts the decoder's outputs using black-box static-collision feedback. We use ``RL fine-tuning'' as shorthand for RL-inspired zeroth-order reward fine-tuning; the update does not use policy likelihoods or policy gradients. We seek to maximize expected output reward:
\begin{equation}
    J(\theta) = \mathbb{E}_{s \sim \rho,\; x_0 \sim \mathcal{A}}\!\left[R\!\left(s,\, T_\theta(x_0, s)\right)\right]
    \label{eq:rl_objective}
\end{equation}
where $\rho$ is the scene distribution and $R$ is a black-box reward. Reward differences under anchor-guided output perturbations provide the learning signal. Its expectation is the gradient of a query-local smoothed surrogate (Eq.~\eqref{eq:local_surrogate}), providing a surrogate update direction for optimizing $J$.

Sampling from the full vocabulary extends reward feedback beyond Stage~1's GT-neighbor pairs, and neighboring vocabulary trajectories provide multiple exploration directions even when the decoder generates only one trajectory. The resulting signal is backpropagated through the trajectory output to model parameters, without differentiating the reward function.

\paragraph{Stage~3 neighbor retrieval.}
For the current model output $a_i$, compute Euclidean distances to the full vocabulary. Excluding zero-distance entries gives $\mathcal A_i^+=\{b\in\mathcal A:\|b-a_i\|_2>0\}$. Retrieve its nearest $N$ anchors:
\begin{equation}
    \mathcal N(a_i)=\operatorname{TopNearest}_{N}(\mathcal A_i^+;a_i),
    \qquad N=\min(N_{\rm ZO},|\mathcal A_i^+|).
    \label{eq:retrieval}
\end{equation}

\paragraph{Step 1: Direction Retrieval and Sampling.} For a sampled starting anchor $x_{0,i}$, generate $a_i=T_\theta(x_{0,i},s)$ and retrieve $N$ nearest anchors $b_{ij}\ne a_i$ using Eq.~\eqref{eq:retrieval}. Construct

\vspace{-1.0\baselineskip}
\begin{equation}
 \sigma_i=\frac1N\sum_{j=1}^N\|b_{ij}-a_i\|_2,\quad
 u_{ij}=\frac{b_{ij}-a_i}{\|b_{ij}-a_i\|_2},\quad
 z_{ij}\overset{\mathrm{i.i.d.}}{\sim}\mathcal N(0,1),\quad
 \widetilde a_{ij}=a_i+\sigma_i z_{ij}u_{ij}.
 \label{eq:anchor_radial_sampling}
\end{equation}
The mean neighbor distance $\sigma_i$ sets the perturbation scale, while the Gaussian scalar $z_{ij}$ samples both signs of each neighbor direction. Neighbors, directions, and scale remain fixed within each estimate.

\paragraph{Step 2: Reward Querying.} Evaluate $r_i=R(s,a_i)$ and $\widetilde r_{ij}=R(s,\widetilde a_{ij})$. Higher reward is better:
\begin{equation}
 R(s,a)=
 \begin{cases}
 \min\{j\in\{0,\ldots,T-1\}:\text{collision at waypoint }j\},
     &\text{if a collision occurs},\\
 T+1, &\text{otherwise},
 \end{cases}
 \label{eq:reward}
\end{equation}
Collision checks cover static obstacles and road boundaries.

\paragraph{Step 3: Directional Signal Estimation.}

\begin{equation}
 \hat g_i=\frac{1}{N\sigma_i}\sum_{j=1}^N
 [R(s,\widetilde a_{ij})-R(s,a_i)]z_{ij}u_{ij}.
 \label{eq:zo_estimator}
\end{equation}
\paragraph{Step 4: Policy Update.} Treating $\hat{g}_i$ as a stop-gradient constant, the RL loss and combined training objective are:
\begin{equation}
    \mathcal{L}_{\text{RL}}(\theta) = -\frac{1}{|\mathcal{B}|}\sum_{i\in\mathcal{B}} \text{sg}(\hat{g}_i)^\top a_i, \qquad
    \mathcal{L}(\theta) = \mathcal{L}_{\text{FM}}(\theta) + \beta\,\mathcal{L}_{\text{RL}}(\theta)
    \label{eq:policy_update}
\end{equation}
where $\mathcal{L}_{\text{FM}}$ combines the residual regression term $\mathcal{L}_{\rm res}$ with the kinematic penalties (Appendix~\ref{app:loss_scaling}).
The update uses a finite-difference learning signal, as in zeroth-order methods such as SPSA~\citep{spall1992multivariate} and ES~\citep{salimans2017evolution}. Here, perturbations are constructed in trajectory space using vocabulary directions. The weight $\beta$ balances reward learning with the FM training objective.

\paragraph{Local interpretation of the update.}
For a fixed query, scene, and set of directions, let $\alpha$ shift the mean of the one-dimensional radial smoothing distribution. The smoothed reward and the expected estimator satisfy

\vspace{-1.5\baselineskip}
\begin{equation}
 \mathcal Q_{ij,\sigma_i}(\alpha)=\mathbb E_z R(s,a_i+(\alpha+\sigma_i z)u_{ij}),\qquad
 \mathbb E[\hat g_i\mid s,a_i,\{u_{ij}\},\sigma_i]=\frac1N\sum_{j=1}^N \mathcal Q'_{ij,\sigma_i}(0)u_{ij}.
 \label{eq:gi_expectation}
\end{equation}
The prime denotes differentiation with respect to $\alpha$, evaluated at $\alpha=0$, so the perturbation distribution remains centered at the current trajectory $a_i$.
This identity holds for bounded measurable rewards, including the piecewise-constant collision reward. It relates the expected signal to derivatives of directionally smoothed rewards. Appendix~\ref{app:gi_expectation} gives the proof, variance bound, and connection to parameter updates.

\paragraph{Local reward objective.}
Fixing the current output and its neighbor directions, define the locally smoothed reward:

\vspace{-1.5\baselineskip}
\begin{equation}
 \resizebox{0.88\linewidth}{!}{$\displaystyle H_i(\Delta)=\frac1N\sum_{j=1}^N \mathbb E_z R\!\left(s,a_i+(u_{ij}^\top\Delta+\sigma_i z)u_{ij}\right),\quad \nabla_\Delta H_i(0)=\mathbb E[\hat g_i\mid s,a_i,\{u_{ij}\},\sigma_i].$}
 \label{eq:local_surrogate}
\end{equation}
$H_i$ averages rewards under directional exploration with shifted radial means, rather than evaluating the full trajectory $a_i+\Delta$. When $\nabla_\Delta H_i(0)\ne0$, a sufficiently small step along this exact gradient increases the fixed local surrogate. This differs from ZPG~\citep{zhang2025zerothorder}, which targets the policy gradient of the global objective: our estimator is unbiased for the gradient of the query-dependent surrogate $H_i$, and its expected signal lies in the span of the $N$ retrieved directions. For sufficiently smooth rewards and small perturbations, this signal approximates a directionally weighted reward gradient; its connection to $\nabla J$ requires the conditions in Appendix~\ref{app:gi_expectation}.

\Needspace{4\baselineskip}
\subsection{Training Pipeline and Inference}
\label{sec:inference}
For EF-initialized configurations, the decoding start $x_k^{(0)}$ denotes the EF-refined trajectory; for the original-prior configuration, it is the raw vocabulary anchor. The FM call count is an inference-time budget: every call reuses the same decoder at $t=0$, so latency can be traded for correction quality without retraining (Section~\ref{sec:guidance_calls}).

All stages retain FM regression. Stage~2 jointly trains EF and FM; Stage~3 freezes EF-specific weights $\phi$ and adds reward learning to FM. The encoder and vocabulary remain fixed. The FM parameters $\theta$ include the shared trajectory projector, so Stage~3 can still change EF's input--output mapping through this projector. Algorithm~\ref{alg:fmrl} summarizes training.

\paragraph{Internal inference and candidate selection.} We decode all $M=2{,}398$ anchors in parallel. For initial trajectory $x_k^{(0)}$, let $x_k^{(1)}=T_\theta(x_k^{(0)},s)$ and $\Delta_k^{(2)}=F_\theta(x_k^{(1)},0,s)$. We retain the $K_{\rm sel}=50$ candidates with the smallest $\|\Delta_k^{(2)}\|_2$. This residual measures how much the decoder would change its first-call output, indicating proximity to a fixed point rather than collision safety; screening therefore costs at most one extra FM forward pass and no reward queries. The downstream selector rejects candidates that fail kinematic, static-collision, or dynamic time-to-collision (TTC) checks, then ranks survivors by safety, efficiency, and comfort. In the production-release configuration, a rule-based planner takes over if none pass. The real-vehicle demonstration configuration reported in Section~\ref{sec:vehicle} instead decodes six sampled anchors without residual screening and does not include this rule-based fallback planner. All internal metrics in Sections~\ref{sec:internal_studies}--\ref{sec:screening} are computed on candidates before this selector, over either the full pool or the residual-screened Top-$K_{\rm sel}$ subset, and they involve static geometry only; the public nuPlan score in Section~\ref{sec:public_scope} instead evaluates the single executed plan in closed loop.

\Needspace{7\baselineskip}

\section{Experiments}
\label{sec:experiments}

We first evaluate the complete planner on nuPlan and demonstrate real-vehicle deployment, then use controlled internal studies to examine Stage~3, EF initialization, decoding calls, and candidate screening; further ablations are in the appendix.

\subsection{Experimental Setup}
\label{sec:experimental_setup}

\paragraph{Public evaluation and deployment.}
We evaluate nuPlan~\citep{caesar2021nuplan} Val14 and Test14-hard using a sampled subset of scenarios per benchmark and simulation mode. Methods within each paired comparison share scenario lists, and all use official closed-loop scoring. Non-reactive (NR) simulation replays logged traffic; reactive (R) simulation uses IDM-driven vehicles. The nuPlan model is trained with a vocabulary of 3,000 anchor trajectories, each represented by 40 waypoints at 10\,Hz including the current state. It generates 3,000 candidates with two EF calls and two FM calls (EF$\times2$, FM$\times2$), then selects one execution trajectory using the score head. We also test the generator on urban roads and measure its runtime on NVIDIA Drive Orin. Appendix~\ref{app:nuplan_protocol} details the public evaluation protocol and scenario sampling.

\paragraph{Internal data and training.}
The main internal Stage~1, Stage~2, and Stage~3 experiments use 120M, 7.5M, and 2.5M samples, respectively, with disjoint training and evaluation cases. The main candidate evaluation uses 10,000 test scenes, with 2,398 trajectories decoded per scene. The ZO sampling and vocabulary ablations (Tables~\ref{tab:zo_exploration} and~\ref{tab:anchor_sweep}) use a separate intersection dataset. Of its approximately 100,000 scenes, 1,000 are reserved for testing.

\paragraph{Metrics and comparisons.}
The nuPlan closed-loop score evaluates the executed plan on a 0--100 scale (higher is better), combining collision safety, drivable-area and driving-direction compliance, route progress, TTC, speed-limit compliance, and comfort (Appendix~\ref{app:nuplan_protocol}). Internal candidate analyses evaluate trajectories before downstream selection. For trajectory accuracy in meters, we use
\begin{equation}
    \text{ADE} = \frac{1}{H}\sum_{t=1}^{H}\|\hat{a}_t - a^*_t\|_2, \qquad \text{FDE} = \|\hat{a}_H - a^*_H\|_2
    \label{eq:ade_fde}
\end{equation}
minADE/FDE is the minimum over the candidate set, evaluated at 3\,s and 8\,s. Static collision rates cover obstacles and road boundaries over the first 40 or all 80 waypoints ($R<40$ or $R<80$); mean reward averages Eq.~\eqref{eq:reward}. EF satisfaction measures the fraction meeting Eq.~\eqref{eq:ef_label}. FM$\times n$ denotes $n$ residual-correction calls; FMRL uses the reward-fine-tuned decoder. We fix the FM call count when comparing training stages and report EF initialization separately.

\paragraph{RL/ZO comparison setup.}
We compare the deterministic FM update with trajectory-space GRPO/DPO-inspired surrogates and standard ZO under matched query budgets; implementation details are given in Appendices~\ref{app:baseline_adaptations} and~\ref{app:zo_comparison}, and results in Appendix~\ref{app:ablation}.

\Needspace{4\baselineskip}
\subsection{Closed-Loop Evaluation}
\label{sec:public_scope}

\paragraph{EF and Stage~3 in closed loop.}
With the score head and downstream processing fixed, enabling EF (DriveAnchor BASE) raises the mean closed-loop score of the executed plans from 72.46 to 74.27 (Table~\ref{tab:nuplan_val14}). Fine-tuning only the decoder in Stage~3 (+RL) raises the mean to 74.74. For context, the paired-subset scores of DriveAnchor BASE exceed those of the reference planners in three of the four settings and overall, while PlanTF scores higher on non-reactive Test14-hard (69.70 vs.\ 64.01); the reference planners are evaluated on the original 100-scenario lists. In a separate evaluation of 21,082 internal closed-loop cases, reward fine-tuning yields 215 net additional passes. Nudging and lane changing show the largest pass-rate gains among the reported categories (Appendix~\ref{app:internal_regression}).

\paragraph{Lane changing.}
Across the 25 lane-changing cases in Table~\ref{tab:nuplan_lanechange}, EF raises the mean score by 4.89 points, with the largest gain on reactive Test14-hard (+15.52). Stage~3 adds another 1.42 points on average, with its largest gain on non-reactive Test14-hard. When road tests reveal a rare lane-change failure, collecting enough corrective demonstrations takes time. DriveAnchor gives developers two ways to address such a failure. They can adjust the corridor request supplied to EF without retraining the model, or use reward feedback on a small set of existing logs to fine-tune the decoder in Stage~3.

\begin{table}[htbp]
\centering
\caption{nuPlan closed-loop scores. DP: Diffusion Planner; BASE: DriveAnchor with EF; DP: public is the official pretrained checkpoint, DP: limited-data is trained on the same frames as our initial FM decoder, and SFT is supervised fine-tuning; +RL: our Stage~3 anchor-guided ZO update. Shading marks our method in the top block and our Stage~3 update in the lower blocks. Top block: reference planners on the original 100-scenario lists per setting, with Overall as the mean of the four settings; the DriveAnchor BASE row repeats its paired result and pair-weighted Overall from the bottom block for reference, so it is evaluated on different scenario lists. Lower blocks: 393 paired scenario--mode pairs (Appendix~\ref{app:public_training}), with Overall as the pair-weighted mean; differences in the text use unrounded values. Bold: best within each block. $^{*}$: input includes a pre-searched reference line or additional proposals. --: not reported, not evaluated, or a setting is missing.}
\label{tab:nuplan_val14}
\small
\begin{tabular}{lrrrrr}
\toprule
\rowcolor{TabHead}& \multicolumn{2}{c}{Val14} & \multicolumn{2}{c}{Test14-hard} & \\
\cmidrule(lr){2-3}\cmidrule(lr){4-5}
\rowcolor{TabHead}Method & NR $\uparrow$ & R $\uparrow$ & NR $\uparrow$ & R $\uparrow$ & Overall \\
\midrule
PlanTF~\citep{cheng2024plantf} & 84.72 & 76.95 & \textbf{69.70} & 61.61 & 73.25 \\
UrbanDriver~\citep{scheel2022urbandriver} & 68.57 & 64.11 & 50.40 & 49.95 & 58.26 \\
STR2-CPKS-800M w/o refine.$^{*}$ & 65.16 & -- & 52.57 & -- & -- \\
PDM-Open$^{*}$~\citep{dauner2023parting} & 53.53 & 54.24 & 33.51 & 35.83 & 44.28 \\
Diffusion-ES w/o LLM~\citep{yang2024diffusiones} & 50.00 & -- & -- & -- & -- \\
GameFormer w/o refine.~\citep{huang2023gameformer} & 13.32 & 8.69 & 7.08 & 6.69 & 8.95 \\
\rowcolor{TabOurs}DriveAnchor BASE (ours) & \textbf{84.89} & \textbf{82.14} & 64.01 & \textbf{65.95} & \textbf{74.27} \\
\midrule
DP: limited-data & 80.07 & 78.77 & 58.50 & 61.05 & 69.62 \\
DP: public & \textbf{92.23} & 79.44 & 73.54 & 68.75 & 78.53 \\
DP + SFT & 89.51 & \textbf{79.52} & 73.63 & \textbf{70.10} & 78.23 \\
\rowcolor{TabOurs}DP + RL (ours) & \textbf{92.23} & 79.44 & \textbf{74.36} & 68.93 & \textbf{78.79} \\
\midrule
DriveAnchor w/o EF & 81.34 & 79.94 & 64.13 & 64.35 & 72.46 \\
DriveAnchor BASE & \textbf{84.89} & \textbf{82.14} & 64.01 & 65.95 & 74.27 \\
\rowcolor{TabOurs}DriveAnchor + RL (ours) & 83.11 & 81.98 & \textbf{66.90} & \textbf{66.89} & \textbf{74.74} \\
\bottomrule
\end{tabular}
\end{table}

\begin{table}[htbp]
\centering
\caption{Lane-changing closed-loop scores for the paired subset of Table~\ref{tab:nuplan_val14}. Differences quoted in the text use unrounded values. Bold marks the best result within each group; shaded rows apply our Stage~3 reward update.}
\label{tab:nuplan_lanechange}
\small
\begin{tabular}{lrrrrr}
\toprule
\rowcolor{TabHead}& \multicolumn{2}{c}{Val14} & \multicolumn{2}{c}{Test14-hard} & \\
\cmidrule(lr){2-3}\cmidrule(lr){4-5}
\rowcolor{TabHead}Method & NR & R & NR & R & Overall \\
\midrule
DP: limited-data & \textbf{61.82} & \textbf{94.70} & 54.04 & 67.29 & 68.97 \\
DP: public & 59.62 & 79.78 & 77.44 & 81.49 & 77.16 \\
DP + SFT & 31.11 & 81.87 & 76.99 & 81.12 & 73.98 \\
\rowcolor{TabOurs}DP + RL (ours) & 59.62 & 79.79 & \textbf{77.45} & \textbf{81.52} & \textbf{77.17} \\
\midrule
DriveAnchor w/o EF & \textbf{96.47} & 90.30 & 76.55 & 60.92 & 77.24 \\
DriveAnchor BASE & 96.42 & 92.33 & 74.81 & 76.44 & 82.13 \\
\rowcolor{TabOurs}DriveAnchor + RL (ours) & 95.86 & \textbf{93.25} & \textbf{78.64} & \textbf{76.55} & \textbf{83.54} \\
\bottomrule
\end{tabular}
\end{table}

\paragraph{Limited-data Diffusion Planner.}
We train the base DriveAnchor FM decoder for two epochs and Diffusion Planner (DP)~\citep{zheng2025diffusionplanner} for three epochs on the same initial frames. With its EF and score head trained, DriveAnchor BASE scores higher in all four settings, with an overall lead of 4.65 points (Table~\ref{tab:nuplan_val14}).

\paragraph{Adapting pretrained Diffusion Planner.}
The official DP checkpoint achieves a higher overall score than DriveAnchor BASE. Starting from that checkpoint, we compare supervised fine-tuning (SFT) with our anchor-guided Stage~3 update. Both fine-tune the decoder for two epochs on the same logs and retain DP's native inference pipeline (Appendix~\ref{app:public_training}). Stage~3 scores 0.25 points above the original checkpoint and 0.56 above SFT overall. The largest gain over the checkpoint is on non-reactive Test14-hard (+0.83 points). On lane-changing cases, the mean scores are 77.16 for the original checkpoint, 73.98 for SFT, and 77.17 for Stage~3.

\Needspace{4\baselineskip}
\subsection{Real-Vehicle Deployment}
\label{sec:vehicle}

\begin{figure}[htbp]
    \centering
    \includegraphics[width=\linewidth]{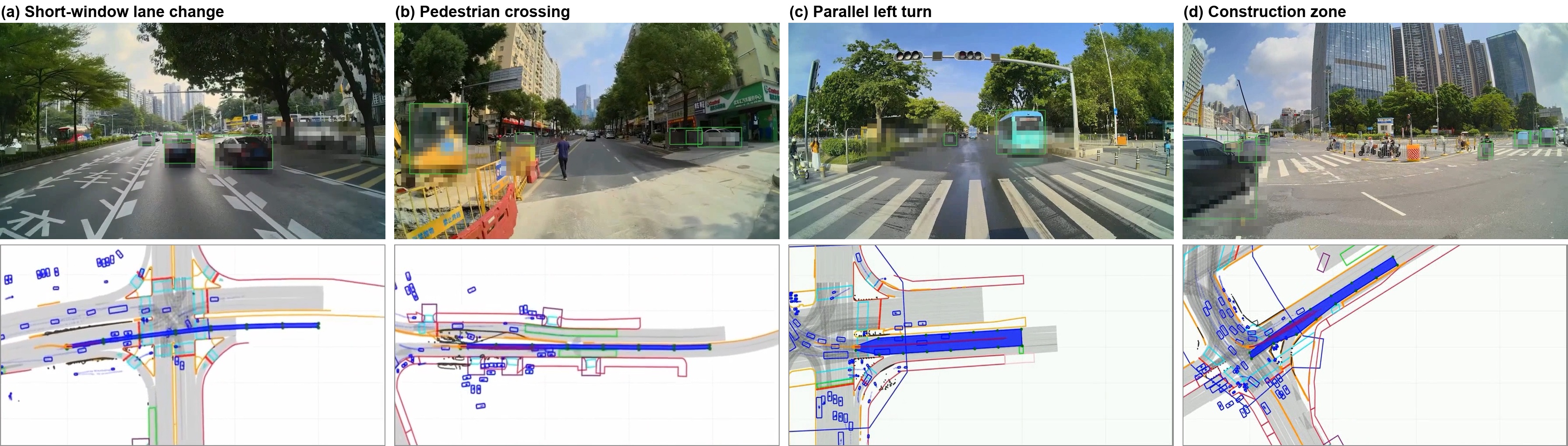}
    \caption{Four anonymized real-vehicle deployment scenarios: (a) short-window lane change, (b) pedestrian crossing, (c) parallel left turn, and (d) construction zone. Top: forward-camera frame. Bottom: the planner's bird's-eye view at the same moment, with the selected trajectory in blue.}
    \label{fig:real_deployment}
\end{figure}

DriveAnchor is integrated into a real vehicle's planning and execution pipeline and tested on urban roads. The system is deployed on thousands of autonomous vehicles operating in urban areas across dozens of cities, with millions of kilometers accumulated and 99\% of the mileage driven autonomously. To balance trajectory diversity and the real-time compute budget, the on-vehicle configuration randomly samples six anchors from the vocabulary, applies EF-guided initialization and FM refinement to each, and passes the six candidates to a rule-based downstream selector, which chooses the trajectory to execute. The reported road tests and accompanying videos use a demonstration configuration without a rule-based fallback planner, distinct from the production-release configuration described in Section~\ref{sec:inference}. Figure~\ref{fig:real_deployment} shows test environments with intersections, pedestrians, two-wheelers, and surrounding vehicles. Deployment configurations, full videos of each scenario, and the complete on-road recording are available on the project page (\projecturl).

\paragraph{On-vehicle runtime.}
In a separate timing measurement on NVIDIA Drive Orin using float16, two EF calls and two FM calls take \textbf{2.06\,ms}/scene on average. This measures the trajectory-generation module, excluding the encoder and downstream selector.

\Needspace{4\baselineskip}
\subsection{Controlled Internal Candidate Analysis}
\label{sec:internal_studies}

Table~\ref{tab:main_results} reports full-candidate results on the 10,000-scenario internal evaluation set. We compare checkpoints at fixed FM call counts to assess Stage~3, then vary EF initialization to examine its effect on candidate quality.

At a fixed two-call FM budget, Stage~3 reduces short-/long-horizon candidate collision from $27.22\%/56.54\%$ to $2.94\%/7.15\%$ and improves minADE@80 from $2.27$ to $1.34$\,m. Adding one EF call to the reward-fine-tuned decoder (FMRL$\times2$ $\to$ EF$\times1$+FMRL$\times2$) improves short-horizon collision ($2.94\%\to1.88\%$) and minADE@80 ($1.34\to0.95$\,m) but increases long-horizon collision ($7.15\%\to8.73\%$), and, on a separate junction set, raises corridor-guidance satisfaction; Section~\ref{sec:guidance_calls} gives the experimental details. This reveals a horizon-dependent accuracy--collision trade-off. Replacing FM with FMRL under EF initialization again reduces both collision rates and minADE.

\begin{table}[htbp]
\centering
\small
\caption{Main internal results on 10,000 randomly selected scenarios. Stage~3 comparisons fix the prior type and FM call count. EF ($^\dagger$) changes the initialization, and additional calls increase inference computation. Collision rates are fractions; bold marks the best result within each prior group.}
\label{tab:main_results}
\setlength{\tabcolsep}{4pt}
\begin{tabular}{lcclccc}
\toprule
\rowcolor{TabHead}& \multicolumn{2}{c}{\textit{Original prior}} & & \multicolumn{3}{c}{\textit{EF-modified prior ($^\dagger$)}} \\
\cmidrule(lr){2-3}\cmidrule(lr){5-7}
\rowcolor{TabHead}\textbf{Metric} & \textbf{FM$\times2$} & \textbf{FMRL$\times2$} & & \shortstack{\textbf{EF$\times1$}\\\textbf{+ FM$\times1$}} & \shortstack{\textbf{EF$\times1$}\\\textbf{+ FMRL$\times1$}} & \shortstack{\textbf{EF$\times1$}\\\textbf{+ FMRL$\times2$}} \\
\midrule
\rowcolor{TabGroup}\multicolumn{7}{l}{\emph{Trajectory Accuracy (m)}} \\
minADE@30 $\downarrow$ & 0.35 & \textbf{0.25} && 0.56 & 0.28 & \textbf{0.22} \\
minFDE@30 $\downarrow$ & 0.80 & \textbf{0.54} && 1.27 & 0.55 & \textbf{0.44} \\
minADE@80 $\downarrow$ & 2.27 & \textbf{1.34} && 2.35 & 1.25 & \textbf{0.95} \\
minFDE@80 $\downarrow$ & 4.42 & \textbf{2.82} && 4.44 & 2.65 & \textbf{1.87} \\
\midrule
\rowcolor{TabGroup}\multicolumn{7}{l}{\emph{Candidate static collision}} \\
Short-horizon collision $\downarrow$ & 0.2722 & \textbf{0.0294} && 0.1313 & 0.0423 & \textbf{0.0188} \\
Long-horizon collision $\downarrow$  & 0.5654 & \textbf{0.0715} && 0.2627 & 0.1178 & \textbf{0.0873} \\
\midrule
\rowcolor{TabGroup}\multicolumn{7}{l}{\emph{Reward}} \\
Mean reward $\uparrow$ & 59.22 & \textbf{78.46} && 70.13 & 77.14 & \textbf{78.87} \\
\bottomrule
\end{tabular}
\end{table}

\Needspace{4\baselineskip}
\subsection{EF Guidance and Decoding-Call Analysis}
\label{sec:guidance_calls}

\paragraph{EF calls and guidance satisfaction.}
On a separate 10,000-scene junction set, each configuration evaluates 2,398 candidates per scene, corresponding to 23,980,000 candidate evaluations. Before Stage~3, one and two EF calls raise guidance satisfaction from 22.10\% to 74.13\% and 87.92\%; with the final checkpoint and two FMRL calls fixed, they raise final-output satisfaction from 62.02\% to 70.31\% and 73.42\%. Thus, repeated EF corrections improve request satisfaction.

\paragraph{FM calls after EF initialization.}
On a separate diagnostic list, each additional FM call after EF initialization further lowers long-horizon static collision at the final Stage~3 checkpoint, with diminishing returns and saturation after three calls (Table~\ref{tab:ef_rl_steps_full}). The full checkpoint-by-call analysis is in Appendix~\ref{app:ablation}.

\Needspace{4\baselineskip}
\subsection{Candidate Screening}
\label{sec:screening}

\paragraph{Residual screening and candidate quality.}
Among the 50 candidates with the smallest residuals, Stage~3 with two FM calls lowers long-horizon collision from 27.80\% to 2.24\% and raises mean reward from 71.05 to 80.48 (Appendix~\ref{app:screening_table}, Table~\ref{tab:top50}).

\section{Conclusion and Limitations}
\label{sec:conclusion}

DriveAnchor connects FM generation, EF control, and zeroth-order reward adaptation through a shared trajectory vocabulary. The same anchors serve as FM priors, EF references, and ZO exploration directions. EF initialization with reward fine-tuning yields the lowest trajectory errors and short-horizon static collision rate (Table~\ref{tab:main_results}). On matched nuPlan scenarios with the score head fixed, EF raises the mean closed-loop score and Stage~3 raises it further. On the official pretrained Diffusion Planner, Stage~3 also scores higher than supervised fine-tuning under the same adaptation budget. We demonstrated a DriveAnchor generator on urban roads in a six-candidate configuration with rule-based trajectory selection. As road tests reveal new long-tail cases, we can continue to adapt the existing planner by adjusting the corridor constraints supplied to EF or by fine-tuning its decoder offline in Stage~3 using a small set of existing logs.

\paragraph{Limitations and future work.}
DriveAnchor relies on a fixed trajectory vocabulary to structure generation and exploration. Its coverage depends on the source data and may be limited for rare maneuvers or unfamiliar road layouts; expanding the vocabulary to cover such cases is a natural extension. The current Stage~3 reward focuses on static obstacles and road boundaries; incorporating dynamic interactions, for example through uncertainty-aware handling of challenging agents~\citep{chen2025uncertainty}, would better align reward adaptation with interactive driving. The ZO analysis characterizes only local optimization behavior under its stated assumptions and does not guarantee global convergence; the global structure of the reward landscape remains to be studied. Our public evaluation uses one sampled scenario subset per benchmark and simulation mode; larger evaluation sets and repeated sampling would better test stability and representativeness, and targeted component adaptation warrants a study of case-repair efficiency. Our Diffusion Planner comparisons cover a same-frame limited-data setting and the adaptation of a public checkpoint; training DriveAnchor at the data scale of public pretrained planners remains to be evaluated. Interaction-rich closed-loop benchmarks such as Bench2Drive~\citep{jia2024bench2drive} would further test whether the gains transfer to interactive decision-making. A matched reward-weighted regression control would further isolate the value of anchor-guided directions, and conditioning neighbor selection and perturbation ranges on ego speed and interaction cues is a natural extension of the exploration design.

\subsubsection*{Acknowledgments}
This work was conducted at Meituan Autonomous Driving. We thank the Meituan Autonomous Driving team for their support and helpful discussions.

\setlength{\bibsep}{3pt plus 1pt minus 1pt}
\bibliographystyle{iclr2027_conference}
\bibliography{references}

\appendix

\section{Evaluation Protocol and Internal Experimental Setup}
\label{app:nuplan_protocol}

\paragraph{nuPlan evaluation protocol.}
We evaluate Val14 and Test14-hard in non-reactive (NR) and reactive (R) closed-loop simulation: NR replays logged traffic, while R uses nuPlan's official IDM configuration. The nuPlan score ranges from 0 to 100 (higher is better) and combines collision safety, drivable-area and driving-direction compliance, route progress, TTC, speed-limit compliance, and comfort. The adapter constructs EF requests only from the navigation route, map boundaries, and current ego state; logged future actors used for offline reward calculation are not inputs to the generator or deployed planner.\footnote{\href{https://github.com/motional/nuplan-devkit/blob/master/nuplan/planning/script/config/simulation/metric_aggregator/closed_loop_reactive_agents_weighted_average.yaml}{Official metric configuration} and \href{https://github.com/motional/nuplan-devkit/blob/master/nuplan/planning/script/config/simulation/observation/idm_agents_observation.yaml}{IDM configuration}.}

\paragraph{Public-model configuration.}
The public model uses 3,000 anchor trajectories, each with 40 two-dimensional waypoints, giving a vocabulary of shape $3{,}000\times40\times2$. The nuPlan BASE configuration uses two EF calls followed by two FM calls. The score head uses anchor geometry, encoded scene context, and a 50-dimensional route-corridor representation to provide scores for selecting a single execution trajectory.

\subsection{Public Training and Adaptation Details}
\label{app:public_training}

\paragraph{Evaluation data.}
We sampled 100 distinct scenarios per setting with a fixed seed for the original baseline evaluation. The paired comparison of DP and DriveAnchor uses 393 scenario--mode pairs (99/98/99/97 for Val14 NR/R and Test14-hard NR/R), and we average the scores over these pairs. The exclusions include one completed Val14 R episode omitted during analysis. The native \texttt{changing\_lane} label identifies 3/6/8/8 cases across the four settings. An asterisk in the table denotes a pre-searched reference line or additional proposals.

\paragraph{DriveAnchor training and deployment.}
We train the base FM decoder on 21,365 nuPlan frames for two epochs. We sample 4,608 of these frames (21.57\%) and use them to train EF for seven epochs and the nuPlan score head for 12 epochs. The EF ablation keeps the FM call count, score head, and downstream processing fixed.

\paragraph{DriveAnchor reward fine-tuning.}
Stage~3 starts from BASE and fine-tunes the FM decoder for two epochs on the same 4,608 frames (9,216 optimizer updates). The encoder, EF, and score head remain frozen. For each scene, four trajectories serve as query centers, each with 16 anchor-guided exploration directions. Evaluating the centers and their perturbations requires 68 reward queries per scene.

The reward checks the 40 waypoints of each trajectory. Its safety term is the index of the first footprint collision or drivable-area violation, or 41 when the trajectory has no violation. A five-point bonus is added when its arc length exceeds 90\% of the logged target's arc length. The internal 8-s reward is defined in Eq.~\eqref{eq:reward}.

An MSE penalty against the frozen BASE model limits changes to its predicted trajectories. The MSE and ZO weights are 1 and 0.01, respectively. We train with AdamW at learning rate $10^{-6}$, zero weight decay, and gradient clipping at 1. Relative parameter $L_2$ change and maximum absolute change are bounded by 0.001. On fixed monitoring samples, waypoint displacement is limited to 0.1\,m RMS and 0.5\,m maximum. Evaluation uses the decoder checkpoint after the second epoch.

\paragraph{Diffusion Planner baselines.}
The limited-data baseline uses the official DP implementation and the same 21,365 frames as initial FM training. It is trained for three epochs with batch size 16 (4,008 optimizer updates), using three layers of hidden dimension 192 and the joint ego/neighbor prediction objective. Its 39 future waypoints at 10\,Hz correspond to DriveAnchor's 40-point representation including the current state. We evaluate the final-epoch EMA checkpoint with the same downstream geometry, speed, risk, and IDM safety rules as DriveAnchor.

The official pretrained DP retains its native 8-s prediction, 80 future waypoints, ten DPM sampling steps, and single ego proposal.

\paragraph{Adapting pretrained DP.}
DP+SFT and DP+RL start from the official checkpoint. Both freeze the encoder and train the decoder on 4,096 frames for two epochs, with effective batch size 16 and 512 optimizer updates. The frames are sampled with fixed seeds from reserved nuPlan logs with complete temporal coverage. Supervision targets and map/agent features are extracted from these logs. Adaptation training, validation, and public evaluation use disjoint recording groups.

DP+RL applies the Stage~3 anchor-guided ZO update using the reward, optimizer, and update bounds described above. Its MSE target is the original model's joint prediction under the same sampling seed; trajectory changes are checked over DP's full prediction horizon. We evaluate the second-epoch checkpoint with native DP inference in FP32, with TF32 disabled. On all 400 episodes, before restricting to the paired subset, the overall scores of the official, SFT, and Stage~3 checkpoints are 78.74, 78.28, and 78.75, respectively.

\subsection{Internal Executed-Plan Regression Evaluation}
\label{app:internal_regression}

Table~\ref{tab:internal_regression} reports the overall change in the number of passing cases and the breakdown for selected scenario categories. Passing cases rise from 9,441 to 9,656: 969 cases change from failure to success and 754 from success to failure, a net gain of 215. The evaluation covers nudging, lane changing, merging, intersection turns, overtaking, yielding, and emergency braking, and measures the plan executed after downstream selection.

\begin{table}[htbp]
\centering
\small
\caption{Internal closed-loop pass rates before and after reward adaptation (\%); changes are in percentage points (pp). Category rows show selected categories and do not sum to the overall row.}
\label{tab:internal_regression}
\begin{tabular}{lrrrrr}
\toprule
\rowcolor{TabHead}Category & Cases & Before & After & $\Delta$\,pp & Additional passes \\
\midrule
\rowcolor{TabOurs}\textbf{Overall} & \textbf{21,082} & \textbf{44.78} & \textbf{45.80} & \textbf{+1.02} & \textbf{+215} \\
Nudge & 5,373 & 41.43 & 44.43 & +3.00 & +161 \\
Lane change & 3,047 & 40.11 & 42.14 & +2.03 & +62 \\
Exchange-area merge & 750 & 46.67 & 47.87 & +1.20 & +9 \\
Intersection turn & 3,084 & 68.81 & 69.07 & +0.26 & +8 \\
Other & 3,290 & 40.33 & 39.42 & $-$0.91 & $-$30 \\
\bottomrule
\end{tabular}
\end{table}

\subsection{Internal Candidate Evaluation Protocol}
\label{app:internal_protocol}

\paragraph{Data provenance.} Stage~1 covers intersections, roundabouts, lane changes, and ramp merges; Stage~2 data are bucketed by dedicated scenario type and cover all EF polygon presets. We construct $\mathcal A$ by FPS on a separate corpus of 100M+ frames collected outside the training and evaluation periods. The stage-wise data counts in Section~\ref{sec:experimental_setup} exclude pretraining of the inherited frozen encoder, whose full data volume is unavailable.

\paragraph{Evaluation and initialization.} The case-disjoint splits and dataset sizes are specified in Section~\ref{sec:experimental_setup}. Table~\ref{tab:main_results} and the cross-maneuver RL comparisons use the main 10,000-scene list; junction-guidance and training-curve diagnostics use separate lists. Tables~\ref{tab:zo_exploration} and~\ref{tab:anchor_sweep} use the smaller intersection-specific dataset and its 1,000 held-out evaluation scenarios, rather than the main evaluation list. ZO statistics aggregate across ranks. The shared initialization is the main Stage~2 EF/FM checkpoint; the $N=16$ reference in Table~\ref{tab:ablation_n} retains the main training and evaluation settings.

\paragraph{Comparison protocol.} Stage~3 comparisons hold the prior and FM call count fixed and optimize the combined FM, kinematic, and reward objective in Eq.~\eqref{eq:policy_update}. Shared upstream projections remain trainable in EF-initialized configurations. Main-table results come from single training runs; reported noise-repetition SDs do not estimate training-seed variability. Appendix~\ref{app:baseline_adaptations} specifies the alternative RL updates.

\paragraph{Metric details.}

Candidate metrics aggregate over the decoded vocabulary and test scenes before downstream selection; residual-screened subsets are reported separately. The horizons use waypoint indices 0--39 and 0--79, respectively. Dynamic-agent collisions are outside the static reward and collision metrics.

\paragraph{Consistency of reward and collision statistics.}
Eq.~\eqref{eq:reward} uses zero-based indexing with $T=80$: a colliding candidate receives the index $j\in\{0,\ldots,79\}$ of its first collision, and a collision-free candidate receives 81. With $p_S=P(R<40)$ and $p_L=P(R<80)$, the mean reward is therefore bounded by $40(p_L-p_S)+81(1-p_L)$ from below and $39p_S+79(p_L-p_S)+81(1-p_L)$ from above. All statistics in Table~\ref{tab:main_results} are computed on the same full candidate pool (10,000 scenes, 2,398 candidates each) before screening and selection, and the reported means are direct averages without normalization. Every mean lies inside its admissible interval (Table~\ref{tab:reward_consistency}).

\begin{table}[htbp]
\centering
\small
\caption{Mean reward in Table~\ref{tab:main_results} and the interval implied by the corresponding short- and long-horizon collision rates.}
\label{tab:reward_consistency}
\begin{tabular}{lcc}
\toprule
\rowcolor{TabHead}Configuration & Admissible interval & Reported mean \\
\midrule
FM$\times2$ & [46.93, 68.98] & 59.22 \\
FMRL$\times2$ & [76.89, 79.68] & 78.46 \\
EF$\times1$+FM$\times1$ & [64.98, 75.22] & 70.13 \\
EF$\times1$+FMRL$\times1$ & [74.48, 79.07] & 77.14 \\
EF$\times1$+FMRL$\times2$ & [76.67, 80.07] & 78.87 \\
\bottomrule
\end{tabular}
\end{table}

\begin{figure}[htbp]
    \centering
    \makebox[\linewidth][c]{\includegraphics[width=\linewidth]{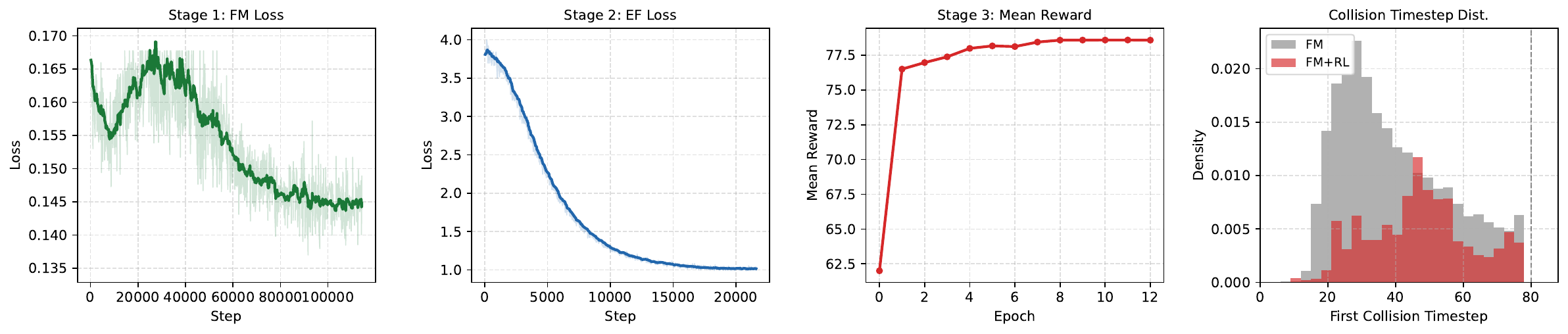}}
    \caption{Training diagnostics. From left to right: Stage~1 FM loss, Stage~2 total loss, Stage~3 mean reward, and first-collision time-step histograms. The reward curve contains 13 saved evaluations, whose indices differ from the epoch labels in Table~\ref{tab:ef_rl_steps_full}. The histograms use a separate 10,000-scene diagnostic set; each model evaluates 2,398 candidates per scene, corresponding to 23,980,000 candidate evaluations, with a common count scale. They show collision indices 0--78, omitting index 79 and the no-collision value 81; index 80 is unused. They do not represent the main evaluation set in Table~\ref{tab:main_results}.}
    \label{fig:training_curves_all}
\end{figure}

\begin{figure}[htbp]
    \centering
    \includegraphics[width=0.92\linewidth]{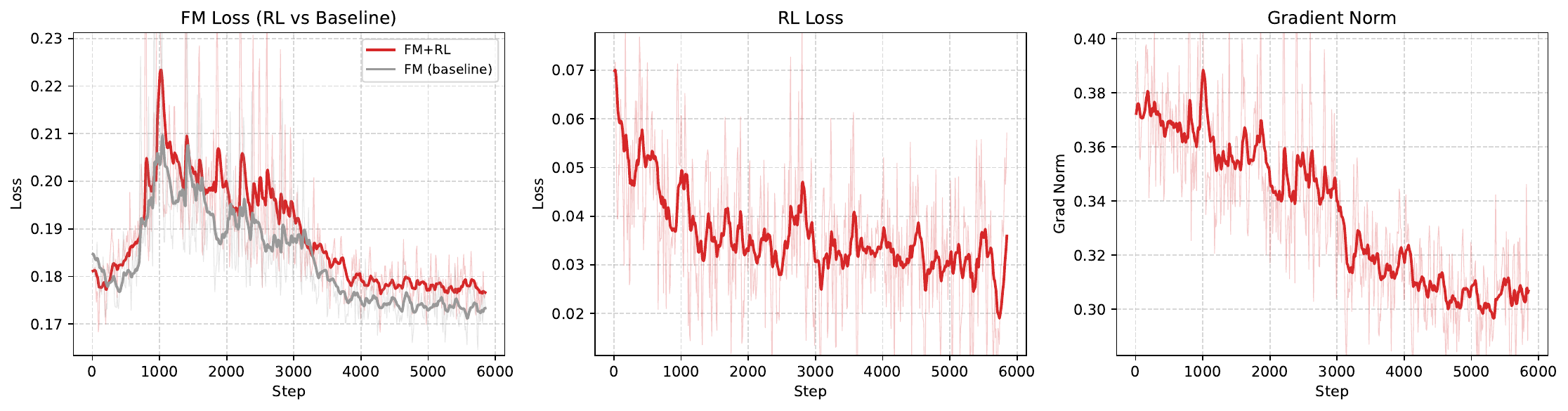}
    \caption{Stage~3 training over 5.8K steps.
    \textbf{Left}: FM regression loss for the RL-fine-tuned model (red) and the frozen pre-Stage~3 baseline (gray); the two remain close.
    \textbf{Center}: RL surrogate loss ($\mathcal{L}_{\text{RL}}$).
    \textbf{Right}: gradient norm.
    Stage~3 combines FM regression with the black-box static collision reward and enables kinematic losses only in the final epoch.}
    \label{fig:app_stage3_rl_loss}
\end{figure}

\paragraph{Training dynamics.}
Figure~\ref{fig:training_curves_all} tracks learning across the three stages. During reward fine-tuning, the FM loss stays close to the frozen baseline, while collision metrics level off late in training. Figure~\ref{fig:app_stage3_rl_loss} shows Stage~3 loss and gradient diagnostics; Appendix~\ref{app:training_curves} provides the remaining training visualizations.

\section{Training Procedure}
\label{app:training_procedure}
In Algorithm~\ref{alg:fmrl}, $\phi$ denotes only EF-specific weights; the trajectory projector is shared by EF and FM (Section~\ref{sec:inference}).

\begin{algorithm}[htbp]
\caption{DriveAnchor training pipeline (internal configuration). Stage~3 applies the query construction to each scene in a batch.}
\label{alg:fmrl}
\begin{algorithmic}[1]
\REQUIRE FM decoder $F_\theta$ and reset map $T_\theta$, Energy Field module $\text{EF}_\phi$, training data $\mathcal{D}$, reward function $R$
\REQUIRE Anchor table $\mathcal{A} = \{a^*_k\}$ (pre-built via FPS from historical corpus), neighbors $N$, RL weight $\beta$
\STATE \textbf{Stage 1 (FM Pretraining):} Train $F_\theta$ on $\mathcal{D}$ with $\mathcal{L}_{\text{FM}}$ \COMMENT{anchor-conditioned residual FM}
\STATE \textbf{Stage 2 (EF Post-training):} Jointly train $F_\theta$ and $\text{EF}_\phi$ on 7.5M Stage~2 training samples with $\mathcal{L}_{\text{FM}} + \mathcal{L}_{\text{EF}}$ (encoder frozen); fix $\phi$ after Stage~2 training \COMMENT{shared frozen features; separate trajectory inputs}
\STATE \textbf{Stage 3 (RL Fine-tuning):} Starting from Stage 2 checkpoint with $\phi$ frozen:
\FOR{each training batch $\mathcal B=\{(s_i,x_{1,i})\}\subset\mathcal D$}
    \STATE For each scene $i$, sample $k_i\sim\text{Uniform}(\{1,\ldots,M\})$, set $x_{0,i}=a^*_{k_i}$, and generate $a_i=T_\theta(x_{0,i},s_i)$ \COMMENT{starts sampled from the full vocabulary}
    \STATE Retrieve $\{b_{ij}\}_{j=1}^N \leftarrow \mathcal{N}(a_i)$ from $\mathcal{A}$ \COMMENT{top-k distance lookup on FM output}
    \STATE Compute the query-adaptive scale and directions, draw independent $z_{ij}$, and generate $\widetilde a_{ij}$ via Eq.~\eqref{eq:anchor_radial_sampling}
    \STATE Query $r_i=R(s_i,a_i)$ and all $\widetilde r_{ij}=R(s_i,\widetilde a_{ij})$ \COMMENT{GridMap collision detection}
    \STATE Compute $\hat{g}_i$ via Eq.~\eqref{eq:zo_estimator} \COMMENT{detach from graph}
    \STATE Compute $\mathcal L_{\mathrm{FM}}$ from local anchor--GT pairs and noisy interpolation (Eqs.~\eqref{eq:training_interp}--\eqref{eq:residual_loss}), with the configured kinematic schedule
    \STATE $\mathcal L\leftarrow\mathcal L_{\mathrm{FM}}(\theta)-\frac{\beta}{|\mathcal B|}\sum_{i\in\mathcal B}\operatorname{sg}(\hat g_i)^\top a_i$ \COMMENT{optimize FM objective; EF-specific $\phi$ frozen}
    \STATE Update $\theta$ using AdamW with gradient $\nabla_\theta\mathcal L$
\ENDFOR
\end{algorithmic}
\end{algorithm}

\paragraph{Regression scaling and kinematic scheduling.}
\label{app:loss_scaling}
The coordinate-mean SmoothL1 loss is $\ell_{\mathrm{SL1}}(p,q)=(2T)^{-1}\sum_{k=1}^{2T}\rho(p_k-q_k)$, where $\rho(e)=e^2/2$ for $|e|<1$ and $\rho(e)=|e|-1/2$ otherwise. The FM objective is $\mathcal L_{\mathrm{FM}}=0.005\,\mathcal L_{\mathrm{res,impl}}+\chi_{\mathrm{kin}}\mathcal L_{\mathrm{kin}}^{\mathrm{FM}}$. Here $\mathcal L_{\mathrm{res,impl}}$ includes the reduction and pairing weights below, and $\chi_{\mathrm{kin}}$ controls kinematic-loss activation. FM regression is retained in all three stages.

The FM and EF heads use a default regression weight of $w_{\mathrm{reg}}=0.005$. Let $T=80$ be the number of trajectory points, distinct from the ZO direction count $N$. Denoting the reduced primary SmoothL1 loss by $\ell_{\mathrm{top1}}$ and the mean neighborhood SmoothL1 loss by $\overline\ell_{50}$, the FM implementation computes
\[
 \mathcal L_{\mathrm{res,impl}}=T(2\ell_{\mathrm{top1}}+\overline\ell_{50}),\qquad
 w_{\mathrm{reg}}\mathcal L_{\mathrm{res,impl}}=0.8\ell_{\mathrm{top1}}+0.4\overline\ell_{50}.
\]
The EF regression contribution is $2T w_{\mathrm{reg}}\ell_{\mathrm{EF}}=0.8\ell_{\mathrm{EF}}$. These coefficients apply to the already-reduced losses, rather than to individual coordinates independently. Kinematic losses are summed with their own per-term coefficients and receive no additional factor of $w_{\mathrm{reg}}$.

In the main Stage~3 experiment, FM kinematic penalties are enabled only in the final epoch: $\chi_{\mathrm{kin}}=0$ in earlier epochs and $\chi_{\mathrm{kin}}=1$ in the final epoch. The RL weight is $\beta=0.2$.

\section{ZO Sampling Configurations}
\label{app:zo_comparison}

\paragraph{Reward queries and budget.}
Eq.~\eqref{eq:reward} uses zero-based waypoint indices: for $T=80$, collision rewards are 0--79 and the collision-free reward is 81. Perturbed outputs are scored directly, without another decoder call or flow-time assignment. The unperturbed reward is shared across directions, requiring $N+1$ queries per output, or $|\mathcal B|(N+1)$ per batch with a common $N$, before additional filtering or resampling. Delaying a collision also increases reward, so we report collision rates alongside mean reward.

For an FM output $x\in\mathbb R^{160}$, let $b_j$ be its $j$th nearest vocabulary anchor and $s_k(x)=k^{-1}\sum_{j=1}^k\|b_j-x\|_2$. Both samplers set their scale from raw L2 distances in trajectory space. The standard ZO implementation recovers this scale by multiplying its internally normalized distances by $\sqrt{T}$, where $T=80$ is the number of trajectory points. Thus, the two methods use the same distance units and scale rule; their scales agree when evaluated at the same query with the same neighbor pool.

\paragraph{Neighbor-scaled standard ZO sampling.}
The standard ZO configuration uses $k=N=16$. With $\lambda=s_{16}(x)$ and independent $\xi_j\sim\mathcal N(0,I_{160})$,
\begin{equation}
 u_j=\xi_j/\|\xi_j\|_2,\qquad \widetilde x_j=x+\lambda u_j.
\end{equation}
The perturbations have fixed length $\lambda$, so $\mathbb E\|\widetilde x_j-x\|_2^2=\lambda^2$.

\paragraph{Anchor-guided sampling.}
Anchor neighbors follow the nonzero-distance convention in Eq.~\eqref{eq:retrieval}.
The anchor-guided configuration uses $k=16$, matching the main experiments. With $\sigma=s_{16}(x)$ and independent $z_j\sim\mathcal N(0,1)$,
\begin{equation}
 u_j=\frac{b_j-x}{\|b_j-x\|_2},\qquad \widetilde x_j=x+\sigma z_j u_j.
\end{equation}
The scalar Gaussian permits exploration in both signs of each anchor direction, with $\mathbb E\|\widetilde x_j-x\|_2^2=\sigma^2$.

Both methods use 17 reward queries per center, including the unperturbed trajectory, without filtering or resampling. Although they share the scale $s_{16}(x)$ at a fixed query with a common neighbor pool, they differ in both direction and radial distribution.

\section{Baseline Adaptations for the FM Decoder}
\label{app:baseline_adaptations}

The following GRPO-inspired and DPO-inspired baselines adapt group-relative updates and preference learning to the anchor-conditioned deterministic FM decoder. They use the surrogate objectives defined below rather than the original likelihood-based objectives. Let $x_i$ be a decoded trajectory, $r_i=R(s,x_i)$ its detached reward, and $\operatorname{sg}$ denote stop-gradient. Gradients do not pass through group construction, pair selection, or reward evaluation.

\paragraph{GRPO-inspired group-relative updates.}
We form candidate groups from seed anchors and nearby vocabulary entries, decode each candidate, and evaluate its reward. We then select the group with the largest reward standard deviation. For its $G$ entries, let $\sigma_r$ and $\bar r$ denote the sample standard deviation and mean reward, respectively. The update is

\begin{equation}
 \begin{aligned}
 A_i &= \frac{r_i-\bar r}{\max(\sigma_r,\varepsilon_r)},\\
 g_i^{\mathrm{grp}} &= \frac{1}{G-1}\sum_{j\ne i}
 \frac{A_j-A_i}{\max(\|x_j-x_i\|_2^2,\varepsilon_d)}(x_j-x_i),\\
 \mathcal L_{\mathrm{grp}} &= -\frac{1}{|\mathcal I|}
 \sum_{i\in\mathcal I}\operatorname{sg}(g_i^{\mathrm{grp}})^\top x_i.
 \end{aligned}
 \label{eq:grpo_inspired_surrogate}
\end{equation}

Here $\varepsilon_r=10^{-4}$ and $\varepsilon_d=10^{-6}$ are numerical floors; groups with $\sigma_r<\varepsilon_r$ contribute zero. $\mathcal I$ includes all entries unless the collision-only option is enabled, in which case it includes only colliding entries. This update retains group-relative reward normalization, using pairwise trajectory directions and a linear surrogate in place of policy-likelihood ratios and clipping. Because it differs from the Gaussian radial estimator used in Stage~3, the expectation identity derived for that estimator does not establish a corresponding identity for this baseline.

\paragraph{DPO-inspired preference regression~\citep{rafailov2023dpo}.}
Among the sampled decoded trajectories, let $\mathcal P$ and $\mathcal N$ denote the collision-free and colliding sets, respectively. We pair each colliding trajectory with its nearest collision-free trajectory. For trajectory dimension $D$, the regression objective is

\begin{equation}
 \begin{aligned}
 j^*(i)&=\arg\min_{j\in\mathcal P}\|x_i-x_j\|_2^2,\\
 d_i&=D^{-1}\|x_i-\operatorname{sg}(x_{j^*(i)})\|_2^2,\\
 \mathcal L_{\mathrm{pref}}&=\frac{1}{|\mathcal N|}
 \sum_{i\in\mathcal N}\operatorname{sg}(w_i)d_i.
 \end{aligned}
 \label{eq:dpo_inspired_surrogate}
\end{equation}

The base weight is $w_i=\min(1,\tau/(d_i+\varepsilon))$, where $\tau$ is a positive scale and $\varepsilon$ is a numerical floor. An optional severity factor assigns larger weights to earlier collisions. The loss is zero when no valid positive--negative pair exists or when the number of negative samples falls below the configured minimum. Both the collision-free targets and the weights are detached. This objective encourages higher-quality outputs through regression; it does not use DPO's reference-policy log-ratio or logistic preference objective.

\section{Network Architecture}
\label{app:architecture}

Table~\ref{tab:architecture} summarizes the modules. The scene encoder is pretrained and frozen throughout all three stages. FM is trained in every stage, while EF is trained in Stage~2. In the internal experiments, the batched trajectory output before screening has shape $[B,2{,}398,160]$, with $B$ scenes and $80\times2$ coordinates per candidate.

\begin{table}[htbp]
\centering
\caption{Network architecture and training hyperparameters for the main internal experiments.}
\label{tab:architecture}
\small
\setlength{\tabcolsep}{5pt}
\begin{tabular}{lll}
\toprule
\rowcolor{TabHead}\textbf{Module} & \textbf{Parameter} & \textbf{Value} \\
\midrule
\rowcolor{TabGroup}\multicolumn{3}{l}{\emph{Shared Encoder (frozen throughout training)}} \\
& Input streams & ego, moving/static obstacles, road graph, traffic light \\
& Transformer layers & 6 \\
& Token dim ($d_{\text{model}}$) & 384 \\
& Attention heads & 6 \\
& Output projection & MLP(384 $\to$ 256) \\
& Output shape & $[B,\,634,\,256]$ \\
\midrule
\rowcolor{TabGroup}\multicolumn{3}{l}{\emph{Anchor Vocabulary \& Shared Projector}} \\
& Vocabulary size $M$ & 2,398 trajectories \\
& Anchor dim & $80 \times 2 = 160$ \\
& Trajectory projector (shared) & MLP(160 $\to$ 256); shared by FM and EF \\
& MLP block & Linear $\to$ LayerNorm $\to$ GELU $\to$ Linear $\to$ Dropout \\
\midrule
\rowcolor{TabGroup}\multicolumn{3}{l}{\emph{FM Decoder (trainable in all stages)}} \\
& Cross-attention heads & 2 \\
& $d_{\text{model}}$ & 256 \\
& Residual head & MLP(256 $\to$ 256) $\to$ MLP(256 $\to$ 160) \\
& FM supervision & Residual regression, target $=x_1-x_t$ (Eq.~\eqref{eq:residual_loss}) \\
& Inference passes & 2 (FM$\times2$) \\
& After screening & $[B,\,K_{\rm sel},\,160]$, with $K_{\rm sel}=50$  \\
\midrule
\rowcolor{TabGroup}\multicolumn{3}{l}{\emph{EF Decoder (Stage~2 trainable)}} \\
& EF input & 50-dim request (scene + 49-dim region) \\
& Scene projector & MLP(1 $\to$ 128) \\
& Region projector & MLP(49 $\to$ 128) \\
& Fusion projector & concat(128, 128, 256) $\to$ MLP(512 $\to$ 256) \\
& Cross-attention heads & 2 \\
& $d_{\text{model}}$ & 256 \\
& EF output head & MLP(256 $\to$ 256) $\to$ MLP(256 $\to$ 160) \\
& EF supervision & SmoothL1; branch on $\operatorname{good}(a_{\rm noisy})$ (Eq.~\eqref{eq:ef_target}) \\
\midrule
\rowcolor{TabGroup}\multicolumn{3}{l}{\emph{Training Hyperparameters}} \\
& Optimizer & AdamW, lr $= 10^{-5}$, batch 128/GPU \\
& FM / EF regression weight & 0.005 \\
& RL loss weight $\beta$ & 0.2 \\
& ZO neighbors $N$ & 16 \\
& RL retrieval & Full-vocabulary top-$N$; no $\epsilon$ filter \\
& Kinematic constraints & speed, accel, jerk, curvature, lat-accel, lat-jerk \\
\bottomrule
\end{tabular}
\end{table}

\section{GridMap-Accelerated Collision Detection}
\label{app:gridmap}

Checking all $N{+}1$ trajectories against every obstacle costs $\mathcal{O}((N{+}1) \cdot T \cdot |\mathcal{O}|)$. We accelerate these checks in two stages. (1) \textbf{Build}: rasterize road boundaries and static objects into a coarse 2D grid ($10\,\text{m/cell}$) using Bresenham traversal, and store the IDs of overlapping obstacles in each cell. (2) \textbf{Query}: project each ego oriented bounding box (OBB) onto the grid, retrieve nearby obstacles, and run precise OBB intersection tests. Once a collision is detected, we skip the remaining waypoints of that trajectory. The grid thus filters obstacle candidates before the geometric checks.

\section{Stage~2 EF Post-training: Kinematic Constraint Configuration}
\label{app:kin_constraints}

\paragraph{Empty reference pools.}When $\mathcal A_{\rm good}$ is empty, satisfactory noisy inputs retain zero targets; the remaining inputs are masked out of EF regression. If all targets are masked, EF regression contributes zero loss while other training objectives remain active.

We use two penalty forms for the six constraints. For upper-bound constraints (speed, acceleration, curvature, and lateral acceleration), the sigmoid-based penalty is zero below the threshold and bounded above. It has a nonzero gradient just above the threshold but a kink at the boundary:
\begin{equation}
    \mathcal{L}_{\text{max}}(x,\, x_{\max}) = \mathbb{E}\!\left[\mathrm{ReLU}\!\left(\operatorname{sigmoid}(x - x_{\max}) - 0.5\right)\right]
    \label{eq:loss_max}
\end{equation}
For smoothness constraints (jerk, lateral jerk), we use a ReLU-squared penalty that strongly penalizes large violations:
\begin{equation}
    \mathcal{L}_{\text{smooth}}(x,\, x_{\max}) = \mathbb{E}\!\left[\max(0,\,|x| - x_{\max})^{2}\right]
    \label{eq:loss_smooth}
\end{equation}
Table~\ref{tab:kin_constraints} details the six losses, their ego-history context, threshold constants, and final loss weights after scaling.

\begin{table}[htbp]
\centering
\caption{Kinematic constraint loss configuration for Stage~2 EF post-training. ``History'' indicates how many ego-history frames are concatenated as input; ``threshold'' is the per-constraint bound; ``weight'' is the product of the base weight and the loss scale coefficient. The penalty forms are defined in Eqs.~\eqref{eq:loss_max} and~\eqref{eq:loss_smooth}.}
\label{tab:kin_constraints}
\small
\setlength{\tabcolsep}{4pt}
\begin{tabular}{lccc}
\toprule
\rowcolor{TabHead}\textbf{Constraint} & \textbf{Ego history} & \textbf{Threshold} & \textbf{Effective weight} \\
\midrule
Speed upper bound          & 1 frame  & $v_{\max}=25.0\,\text{m/s}$               & $1.0 \times 0.1 = 0.1$ \\
Acceleration upper bound   & 2 frames & $a_{\max}=4.5\,\text{m/s}^2$              & $0.01 \times 0.01 = 1\times10^{-4}$ \\
Jerk smoothness            & 3 frames & $j_{\max}=4.5\,\text{m/s}^3$              & $0.001 \times 0.1 = 1\times10^{-4}$ \\
Curvature upper bound      & 2 frames & $\kappa_{\max}=0.4339\,\text{m}^{-1}$     & $0.005 \times 0.1 = 5\times10^{-4}$ \\
Lateral acceleration bound & 2 frames & $a_{\text{lat},\max}=1.0\,\text{m/s}^2$ & $0.01 \times 0.005 = 5\times10^{-5}$ \\
Lateral jerk smoothness    & 3 frames & $j_{\text{lat},\max}=1.0\,\text{m/s}^3$  & $0.001 \times 0.1 = 1\times10^{-4}$ \\
\bottomrule
\end{tabular}
\end{table}

\section{Training Compute}
\label{app:compute}

All three stages of the main internal experiments run on A100-80G GPUs with PyTorch~2.3 and AdamW (lr$=10^{-5}$, weight decay $0.01$, batch size 128/GPU). Stage~1 uses 120M samples, and Stage~2 jointly trains EF and FM on 7.5M samples. Stage~3 performs zeroth-order reward fine-tuning on 2.5M samples using 64 GPUs. Table~\ref{tab:compute} reports elapsed training time and GPU$\cdot$hours for each stage.

\begin{table}[htbp]
\centering
\caption{Training compute per stage for the main internal experiments.}
\label{tab:compute}
\small
\begin{tabular}{lccc}
\toprule
\rowcolor{TabHead}\textbf{Stage} & \textbf{GPUs} & \textbf{Wall time} & \textbf{GPU$\cdot$Hours} \\
\midrule
Stage~1 (FM Pretraining)    & 32$\times$A100-80G & 28.2\,h & 902 \\
Stage~2 (EF Post-training)  & 32$\times$A100-80G & 16.9\,h & 541 \\
Stage~3 (RL Fine-tuning)    & 64$\times$A100-80G &  3.8\,h & 243 \\
\bottomrule
\end{tabular}
\end{table}

\section{Ablation Studies}
\label{app:ablation}

\subsection{Residual-Screened Candidate Quality}
\label{app:screening_table}

The following table gives the full residual-screening comparison referenced in Section~\ref{sec:screening}. The statistics are computed before the downstream selector.

\begin{table}[htbp]
\centering
\caption{Results for the Top-$K_{\rm sel}$ candidates selected in each scene by the second-call residual norm (Section~\ref{sec:inference}), before the downstream selector. Short- and long-horizon collision rates are fractions. Each column corresponds to the same configuration in Table~\ref{tab:main_results}. The first two columns use the original prior; the last three use EF initialization. Bold indicates the best result within each group.}
\label{tab:top50}
\small
\begin{tabular}{lccccc}
\toprule
\rowcolor{TabHead}\textbf{Metric} & \textbf{FM$\times2$} & \textbf{FMRL$\times2$} & \textbf{EF$\times1$+FM$\times1$} & \textbf{EF$\times1$+FMRL$\times1$} & \textbf{EF$\times1$+FMRL$\times2$} \\
\midrule
Short-horizon $\downarrow$ & 0.1373 & \textbf{0.0047} & 0.2276 & 0.0264 & \textbf{0.0062} \\
Long-horizon $\downarrow$  & 0.2780 & \textbf{0.0224} & 0.4024 & 0.0842 & \textbf{0.0441} \\
Mean reward $\uparrow$ & 71.05 & \textbf{80.48} & 63.22 & 78.54 & \textbf{80.14} \\
\bottomrule
\end{tabular}
\end{table}

\begin{figure}[htbp]
    \centering
    \includegraphics[width=0.95\linewidth]{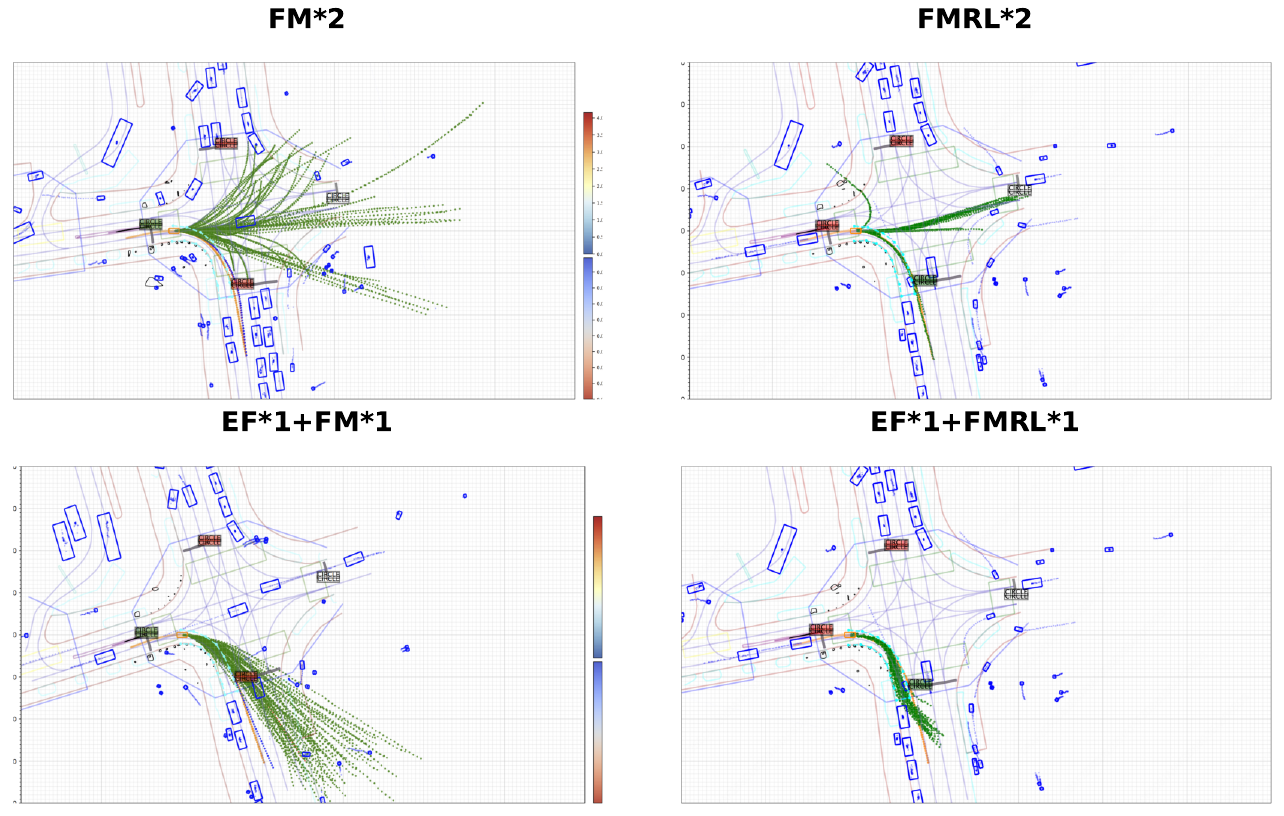}
    \caption{Example trajectories from the residual-selected Top-$K_{\rm sel}$ candidates under four configurations, before the downstream selector. Spatial coordinates are in meters.}
    \label{fig:trajectory_viz}
\end{figure}

\subsection{Stage~3 Transfer across Maneuver Subsets}
\label{sec:ood}

We fine-tune Stage~3 on lane-change (Lac) or junction (Jct) subsets and evaluate transfer to the other subset. Both maneuver types are present in the shared base-stage training data, so transfer here concerns the fine-tuning subset. Results are collision rates after one and two FM calls (\%, mean${\pm}$SD over three repetitions with different noise samples). GRPO-inspired and DPO-inspired use the deterministic FM surrogates in Appendix~\ref{app:baseline_adaptations} rather than the original likelihood objectives.

\begin{table}[htbp]
\centering
\caption{Collision rates after Stage~3 transfer between maneuver subsets on 10,000 evaluation scenarios (\%, mean${\pm}$SD over three repetitions with different noise samples). Baselines use the adaptations described in the text.}
\label{tab:ood}
\small
\begin{tabular}{lccc}
\toprule
\rowcolor{TabHead}Direction & \textbf{ZO (ours)} & \shortstack{GRPO-\\inspired} & \shortstack{DPO-\\inspired} \\
\midrule
Jct$\to$Lac, FM$\times1$ & $\mathbf{4.41{\pm}0.60}$ & $6.65{\pm}0.96$ & $7.04{\pm}0.61$ \\
Jct$\to$Lac, FM$\times2$ & $\mathbf{1.77{\pm}0.55}$ & $3.75{\pm}0.90$ & $3.72{\pm}0.59$ \\
Lac$\to$Jct, FM$\times1$ & $\mathbf{3.49{\pm}0.68}$ & $7.07{\pm}0.90$ & $5.02{\pm}1.05$ \\
Lac$\to$Jct, FM$\times2$ & $1.57{\pm}0.58$ & $3.80{\pm}0.50$ & $\mathbf{1.44{\pm}0.43}$ \\
\bottomrule
\end{tabular}
\end{table}

Among the three compared update rules, ZO achieves the lowest mean collision rate in three of the four combinations of transfer direction and call count. DPO-inspired is lowest for Lac$\to$Jct with two calls ($1.44\%$, compared with $1.57\%$ for ZO). Under the shared base-stage initialization, the strongest transfer result therefore depends on the maneuver direction and decoding budget.

\Needspace{4\baselineskip}
\subsection{Anchor-Guided versus Standard ZO Exploration}

Tables~\ref{tab:zo_exploration} and~\ref{tab:anchor_sweep} report lightweight ablations on a smaller, intersection-specific dataset containing approximately 100,000 scenarios. Evaluation uses 1,000 held-out scenarios excluded from training. These experiments use a shared Stage~2 initialization and a separate reward-fine-tuning setting from the main experiment; their absolute collision rates are therefore not directly comparable to Table~\ref{tab:main_results}.

Table~\ref{tab:zo_exploration} tests whether the complete anchor-guided sampling configuration improves on standard ZO. Both use $N=16$, 17 reward queries per center, and the same neighbor-distance scale rule. Evaluation generates candidates with one EF call followed by one FM call, using FMRL after reward fine-tuning. Anchor-guided ZO improves all four metrics relative to Stage~2 initialization and standard ZO; standard ZO lowers collision but increases trajectory error. This comparison supports the sampling configuration, but does not isolate direction selection: the variants differ in both direction distribution and radial perturbation, with standard ZO using a fixed radius and anchor-guided ZO using a signed Gaussian scalar (Appendix~\ref{app:zo_comparison}).

\begin{table}[H]
\centering
\small
\setlength{\tabcolsep}{4pt}
\caption{ZO exploration comparison on 1,000 held-out scenarios from the smaller intersection-specific dataset, with statistics aggregated across ranks. Candidate generation uses EF$\times1$+FM$\times1$ before fine-tuning and EF$\times1$+FMRL$\times1$ afterward. Both ZO variants start from the shared Stage~2 checkpoint and use $N=16$. Long-horizon static collision rates ($R<80$) are reported for all $2{,}398$ candidates and the Top-50 candidates selected per scene by residual norm, with the additional screening call described in Section~\ref{sec:inference}.}
\label{tab:zo_exploration}
\begin{tabular}{lccc}
\toprule
\rowcolor{TabHead}Metric $\downarrow$ & Before fine-tuning & \shortstack{Standard ZO\\$N=16$} & \shortstack{Anchor ZO\\$N=16$} \\
\midrule
minADE@80 (m) & 1.65 & 1.78 & \textbf{1.50} \\
minFDE@80 (m) & 3.97 & 4.34 & \textbf{3.47} \\
All-candidate collision (\%) & 3.06 & 1.97 & \textbf{0.28} \\
Top-50 collision (\%) & 2.80 & 1.11 & \textbf{0.44} \\
\bottomrule
\end{tabular}
\end{table}

\Needspace{4\baselineskip}
\subsection{Vocabulary and Neighbor-Count Ablations}

\paragraph{Vocabulary size and construction.}
Using EF$\times1$+FMRL$\times1$ on the 1,000 held-out scenarios from the smaller intersection-specific dataset, we vary vocabulary construction (FPS or K-means) and size $M$. Each vocabulary configuration undergoes a separate Stage~3 reward-fine-tuning run from the shared Stage~2 initialization using its corresponding vocabulary; the vocabulary is not merely replaced at evaluation time. Increasing the FPS vocabulary from 100 to 2,398 anchors lowers long-horizon collision from $0.54\%$ to $0.28\%$ (Table~\ref{tab:anchor_sweep}). FPS and K-means perform similarly for $M\geq500$, while K-means is lower at $M=100$; increasing from 2,000 to 2,398 anchors yields a 0.01-percentage-point gain.

\begin{table}[htbp]
\centering
\caption{Vocabulary construction and size ablation on 1,000 held-out scenarios from the smaller intersection-specific dataset with EF$\times1$$+$FMRL$\times1$. Each vocabulary configuration is separately reward-fine-tuned from the shared Stage~2 initialization. Values are long-horizon static candidate collision rates ($R<80$, \%) before the selector, reported as mean${\pm}$SD over three repetitions with different noise samples.}
\label{tab:anchor_sweep}
\small
\begin{tabular}{lccccc}
\toprule
\rowcolor{TabHead}$M$ & 100 & 500 & 1,000 & 2,000 & 2,398 \\
\midrule
\rowcolor{TabOurs}\textbf{FPS (ours)} & $0.54{\pm}0.03$ & $0.38{\pm}0.01$ & $0.33{\pm}0.01$ & $0.29{\pm}0.01$ & $\mathbf{0.28{\pm}0.01}$ \\
K-means & $0.47{\pm}0.02$ & $0.37{\pm}0.01$ & $0.32{\pm}0.01$ & $0.29{\pm}0.01$ & -- \\
\bottomrule
\end{tabular}
\end{table}

\paragraph{Number of anchor neighbors.}
The $N=4$ and $N=8$ runs use a smaller ablation training set, while $N=16$ reuses the main training setting; all configurations are evaluated on the same 10,000 scenes. Increasing $N$ from 4 to 8 lowers short- and long-horizon collision from $27.52\%/56.75\%$ to $3.04\%/7.69\%$ (Table~\ref{tab:ablation_n}). This change also increases the query budget and changes the direction set and distance scale; the $N=16$ main result uses different training settings and is listed separately as a reference.

\begin{table}[!htbp]
\centering
\small
\caption{Ablation on the number of anchor neighbors $N$, initialized from the main experiment's trained Stage~2 checkpoint containing EF and FM. Collision rates are measured on the same 10,000 held-out scenes; arrows compare $N=8$ and $N=4$ trained with the smaller dataset. The $N{=}16$ row reuses the FMRL$\times2$ result from Table~\ref{tab:main_results} as a reference and uses different training data.}
\label{tab:ablation_n}
\begin{tabular}{lccc}
\toprule
\rowcolor{TabHead}$N$ & Short-horizon & Long-horizon & Mean reward \\
\midrule
4           & 0.2752          & 0.5675          & 59.10 \\
8           & 0.0304$\downarrow$ & 0.0769$\downarrow$ & 78.25$\uparrow$ \\
\midrule
\rowcolor{TabOurs}16 (main-table reference) & 0.0294 & 0.0715 & 78.46 \\
\bottomrule
\end{tabular}
\end{table}

\paragraph{Vocabulary sampling and kinematic feasibility.}
We use all 2,398 anchors in the internal trajectory vocabulary as sampling centers. Each anchor contains 80 XY waypoints. For each center $x$, we retrieve its 16 nearest neighbors from the full vocabulary, excluding itself, and construct unit directions $u_j=(b_j-x)/\|b_j-x\|_2$ in flattened trajectory space. We evaluate Gaussian perturbations
\[
    y=x+\lambda z u_j,\qquad z\sim\mathcal N(0,1),
    \qquad \lambda\in\{1,5,10,20\}.
\]
Here $\lambda$ controls the perturbation scale of the entire trajectory; the realized perturbation distance is $\lambda|z|$. All original anchors are included without feasibility filtering.

We evaluate the perturbed trajectories using a time interval $h=0.1$\,s and the following discrete constraints:
\[
    \max_t\left\|\frac{y_{t+1}-y_t}{h}\right\|_2
    \le 25\,\mathrm{m/s},
    \qquad
    \max_t\left\|\frac{y_{t+2}-2y_{t+1}+y_t}{h^2}\right\|_2
    \le 4.5\,\mathrm{m/s^2}.
\]
Only differences within the trajectory are checked. For each center--direction pair, we compute the feasible interval along $y=x+qu_j$ and integrate its Gaussian probability analytically. Table~\ref{tab:sampling_va} reports the mean probability over all $2{,}398\times16$ center--direction pairs.

\begin{table}[htbp]
    \centering
    \small
    \caption{Mean probability of satisfying both discrete speed and acceleration constraints under Gaussian perturbations around all 2,398 anchors, using 16 nearest-neighbor directions per anchor.}
    \label{tab:sampling_va}
    \begin{tabular}{lcccc}
        \toprule
        \rowcolor{TabHead}Perturbation scale $\lambda$ & 1 & 5 & 10 & 20 \\
        \midrule
        Feasibility probability (\%) & 97.43 & 96.81 & 94.38 & 84.85 \\
        \bottomrule
    \end{tabular}
\end{table}

The mean feasibility probability decreases as the perturbation scale increases, from 97.43\% at $\lambda=1$ to 84.85\% at $\lambda=20$. This diagnostic measures discrete speed and acceleration feasibility, not collision avoidance or closed-loop safety.

\Needspace{4\baselineskip}
\subsection{Additional Training Diagnostics}

\paragraph{RL training progress and decoding calls.}
Each configuration applies EF once, then calls FM consecutively with $t=0$ reset at every call. The table reports collision percentages over all generated candidates before Top-$K_{\rm sel}$ filtering. Epochs correspond to checkpoints from one training run, rather than independent training seeds.

\begin{table}[htbp]
\centering
\small
\caption{Long-horizon static collision rates ($R<80$, \%) over all generated candidates on a separate 10,000-scene diagnostic list, distinct from the main evaluation list of Table~\ref{tab:main_results}. Each configuration applies one EF call, followed by the number of FM calls shown in the column heading; post-RL rows use the corresponding FMRL checkpoint.}
\label{tab:ef_rl_steps_full}
\setlength{\tabcolsep}{3pt}
\begin{tabular}{lrrrr@{\hspace{12pt}}lrrrr}
\toprule
\rowcolor{TabHead}\rowcolor{TabHead}Epoch & FM $\times1$ & FM $\times2$ & FM $\times3$ & FM $\times4$ & Epoch & FM $\times1$ & FM $\times2$ & FM $\times3$ & FM $\times4$ \\
\cmidrule(lr){1-5}\cmidrule(lr){6-10}
Before RL & 26.71 & 20.62 & 16.37 & 13.40 & -- & -- & -- & -- & -- \\
0 & 18.19 & 10.36 & 5.69 & 3.60 & 6 & 2.02 & 0.55 & 0.37 & 0.35 \\
1 & 6.72 & 1.26 & 0.57 & 0.43 & 7 & 1.90 & 0.52 & 0.36 & 0.35 \\
2 & 3.54 & 0.88 & 0.52 & 0.41 & 8 & 1.93 & 0.52 & 0.35 & 0.34 \\
3 & 2.96 & 0.78 & 0.54 & 0.44 & 9 & 1.91 & 0.51 & 0.34 & 0.34 \\
4 & 2.60 & 0.70 & 0.41 & 0.36 & 10 & 1.94 & 0.53 & 0.35 & 0.34 \\
5 & 2.27 & 0.67 & 0.42 & 0.36 & 11 & 1.91 & 0.52 & 0.34 & 0.34 \\
\bottomrule
\end{tabular}
\end{table}

\section{Additional Figures}

\Needspace{4\baselineskip}
\subsection{EF Polygon Presets}
\label{app:ef_polygons}

\begin{figure}[!htbp]
    \centering
    \captionsetup[subfigure]{skip=2pt}
    \begin{subfigure}[b]{0.45\linewidth}
        \includegraphics[width=\linewidth]{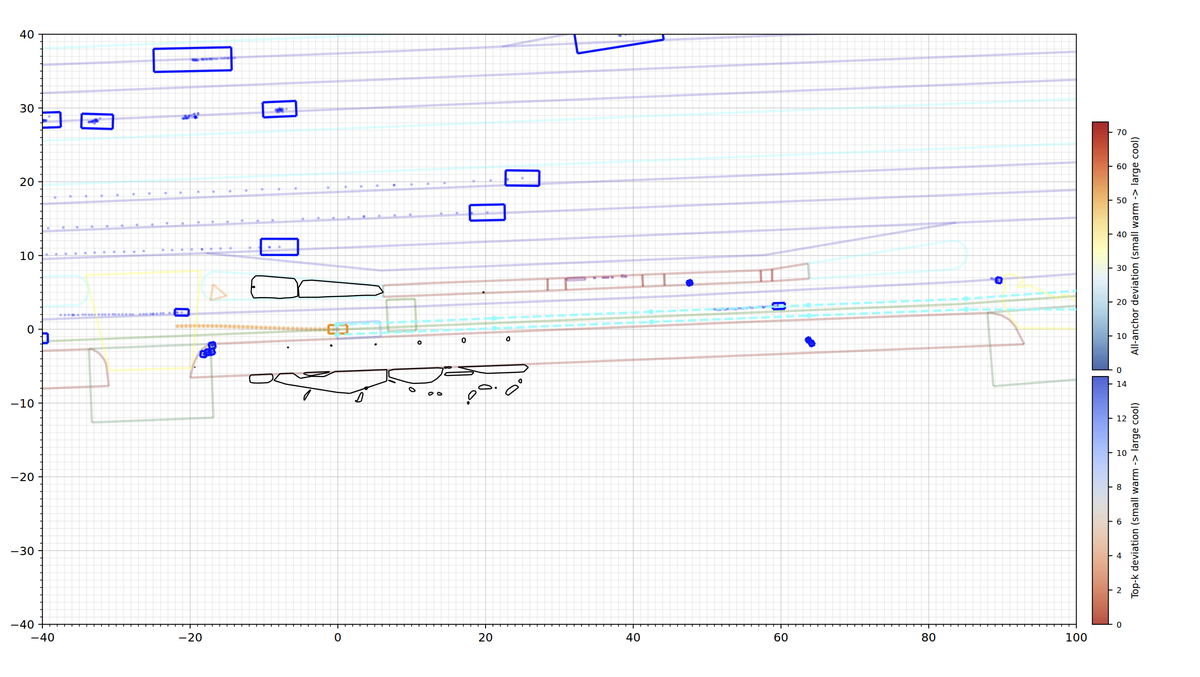}
        \caption{Lane keep (straight)}
    \end{subfigure}
    \hfill
    \begin{subfigure}[b]{0.45\linewidth}
        \includegraphics[width=\linewidth]{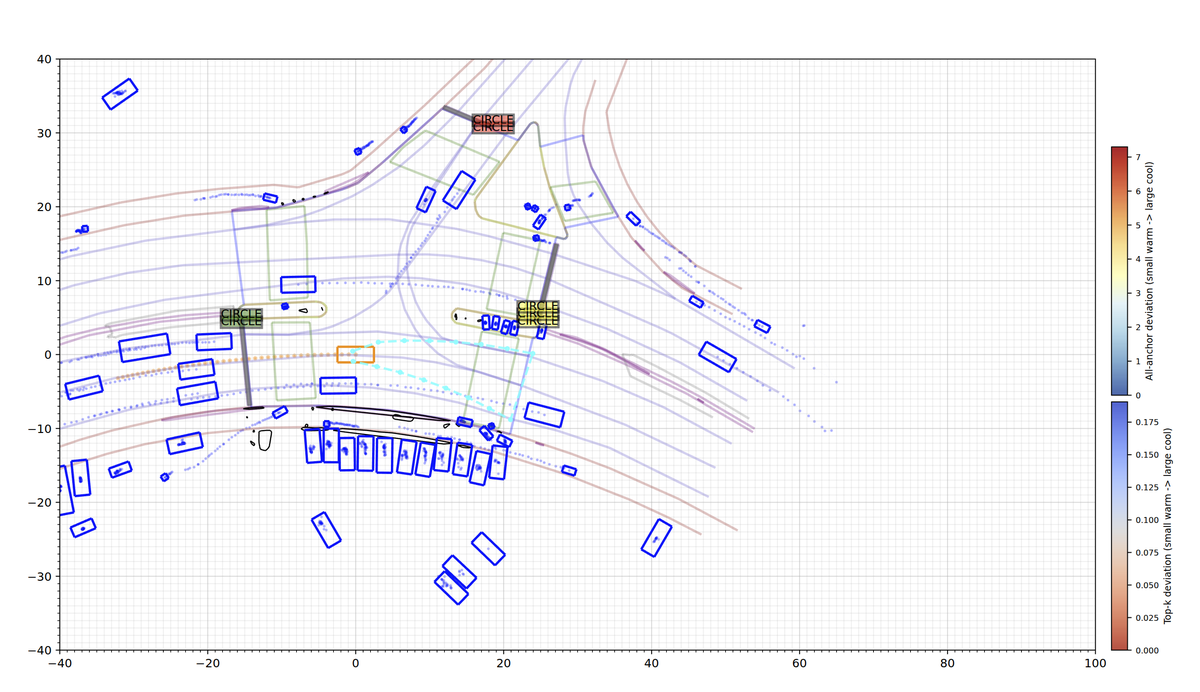}
        \caption{Intersection straight}
    \end{subfigure}
    \\[2pt]
    \begin{subfigure}[b]{0.45\linewidth}
        \includegraphics[width=\linewidth]{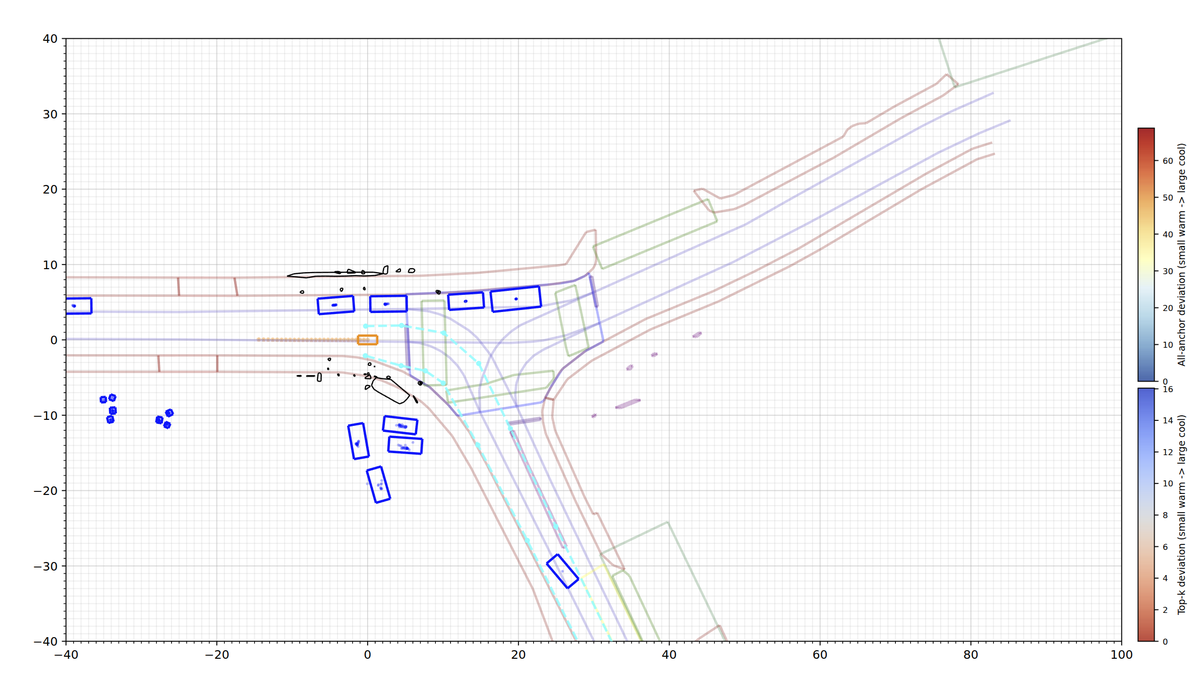}
        \caption{Intersection right turn}
    \end{subfigure}
    \hfill
    \begin{subfigure}[b]{0.45\linewidth}
        \includegraphics[width=\linewidth]{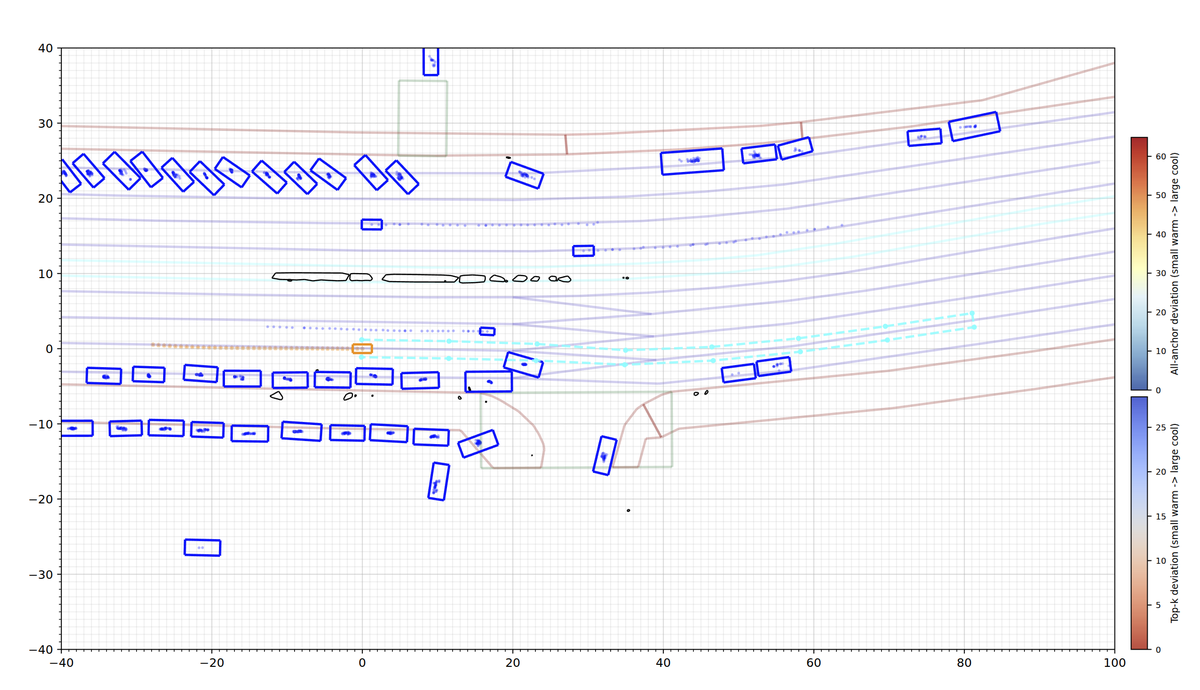}
        \caption{Lane change (current lane)}
    \end{subfigure}
    \\[2pt]
    \begin{subfigure}[b]{0.45\linewidth}
        \includegraphics[width=\linewidth]{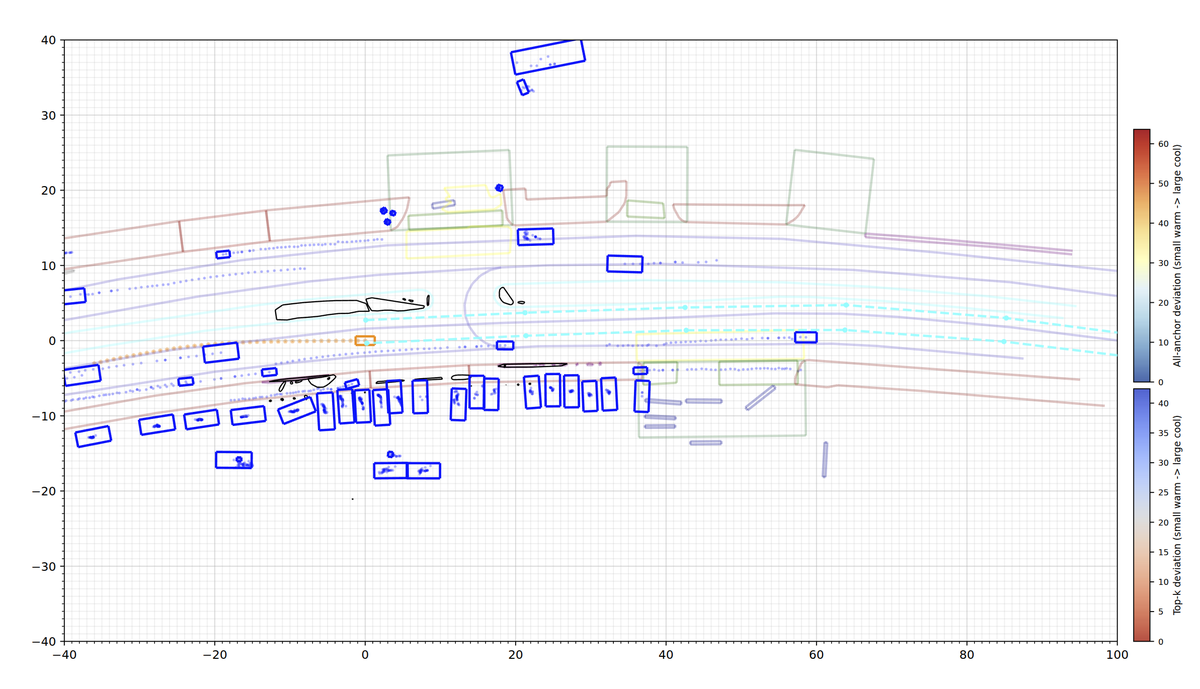}
        \caption{Lane change (target lane)}
    \end{subfigure}
    \hfill
    \begin{subfigure}[b]{0.45\linewidth}
        \includegraphics[width=\linewidth]{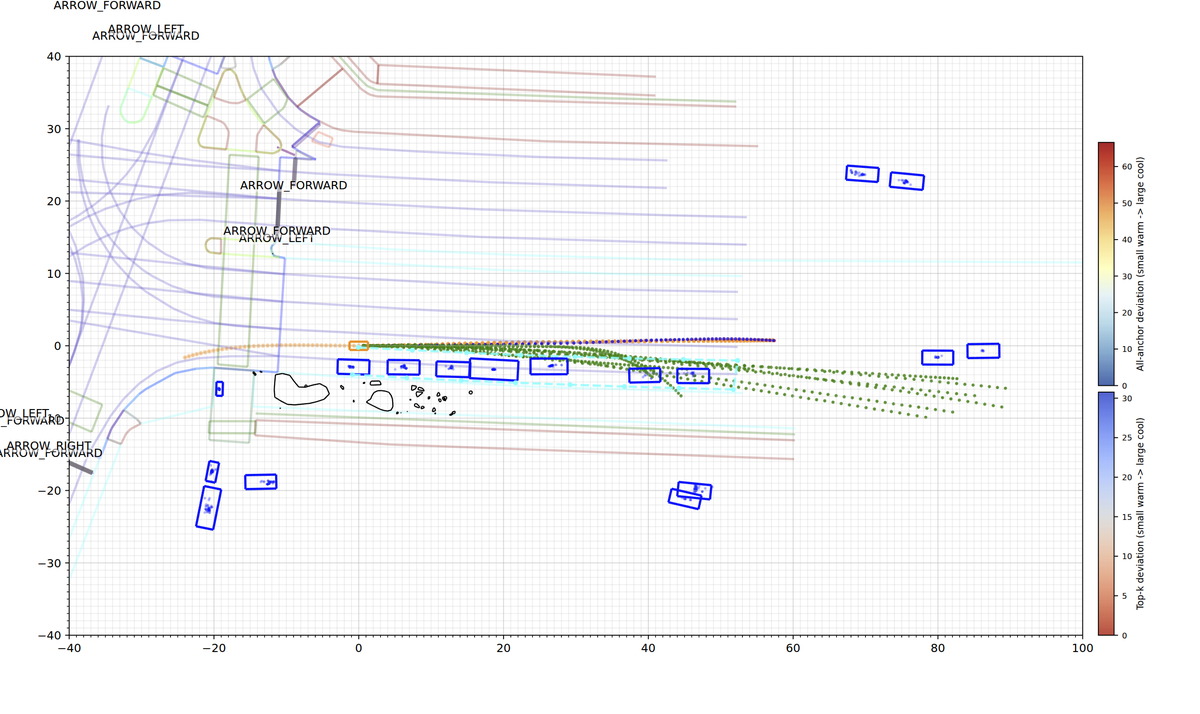}
        \caption{Hard-end short lane change}
    \end{subfigure}
    \caption{EF polygon presets for six driving scenarios. The geometric label checks entry into the corridor and allows at most one exit through the designated edge. An optional label-construction check requires the trajectory to remain inside at a specified time. The hard-end short lane-change preset (bottom right) limits the requested maneuver to a short spatial window. Spatial coordinates are in meters.}
    \label{fig:app_ef_polygons}
\end{figure}

\paragraph{EF request encoding.}
The request is encoded as 50 values: one scene type, 16 interleaved $(x,y,\mathrm{enterable})$ vertex triples, and one region-expansion flag. The 49-dimensional region block, which includes the expansion flag, is projected separately from the scene type. Vertices 0--7 trace the left boundary forward, and vertices 8--15 trace the right boundary backward. The designated forward exit is therefore edge 7, connecting vertices 7 and 8 under zero-based indexing. For a lane-change request that starts outside the polygon, both endpoints of the first entry edge must be marked enterable; a trajectory that starts inside automatically passes this entry check. The feature has no separate exit-index or time scalar. The optional time gate in Eq.~\eqref{eq:ef_label} is applied when constructing labels.

\Needspace{7\baselineskip}
\subsection{Training and Visualization}
\label{app:training_curves}

\begin{figure}[htbp]
    \centering
    \includegraphics[width=0.88\linewidth]{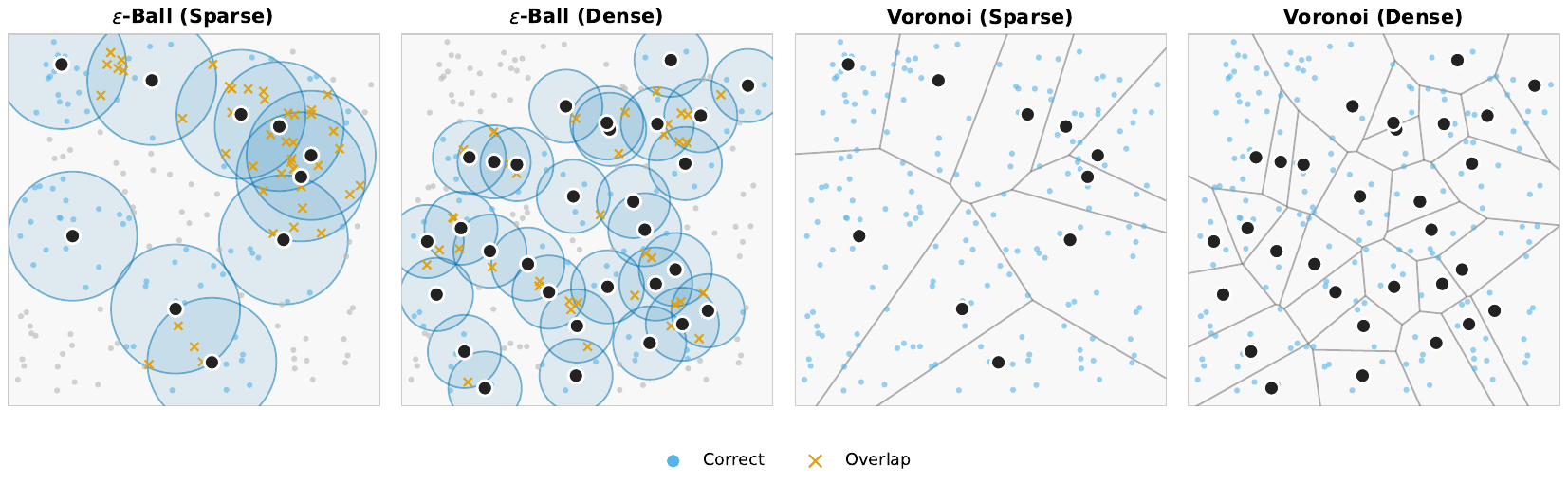}
    \caption{$\epsilon$-ball regions and Voronoi partitions for dense and sparse anchors. Blue points have a unique assignment, orange crosses have multiple assignments, and gray points are uncovered; white-bordered circles mark anchors. $\epsilon$-balls leave overlaps and gaps, whereas nearest-anchor assignment, with consistent tie-breaking, provides complete, non-overlapping coverage in both anchor regimes.}
    \label{fig:voronoi_compare}
\end{figure}

\begin{figure}[htbp]
    \centering
    \begin{subfigure}[t]{0.48\linewidth}
        \includegraphics[width=\linewidth, height=5.5cm, keepaspectratio]{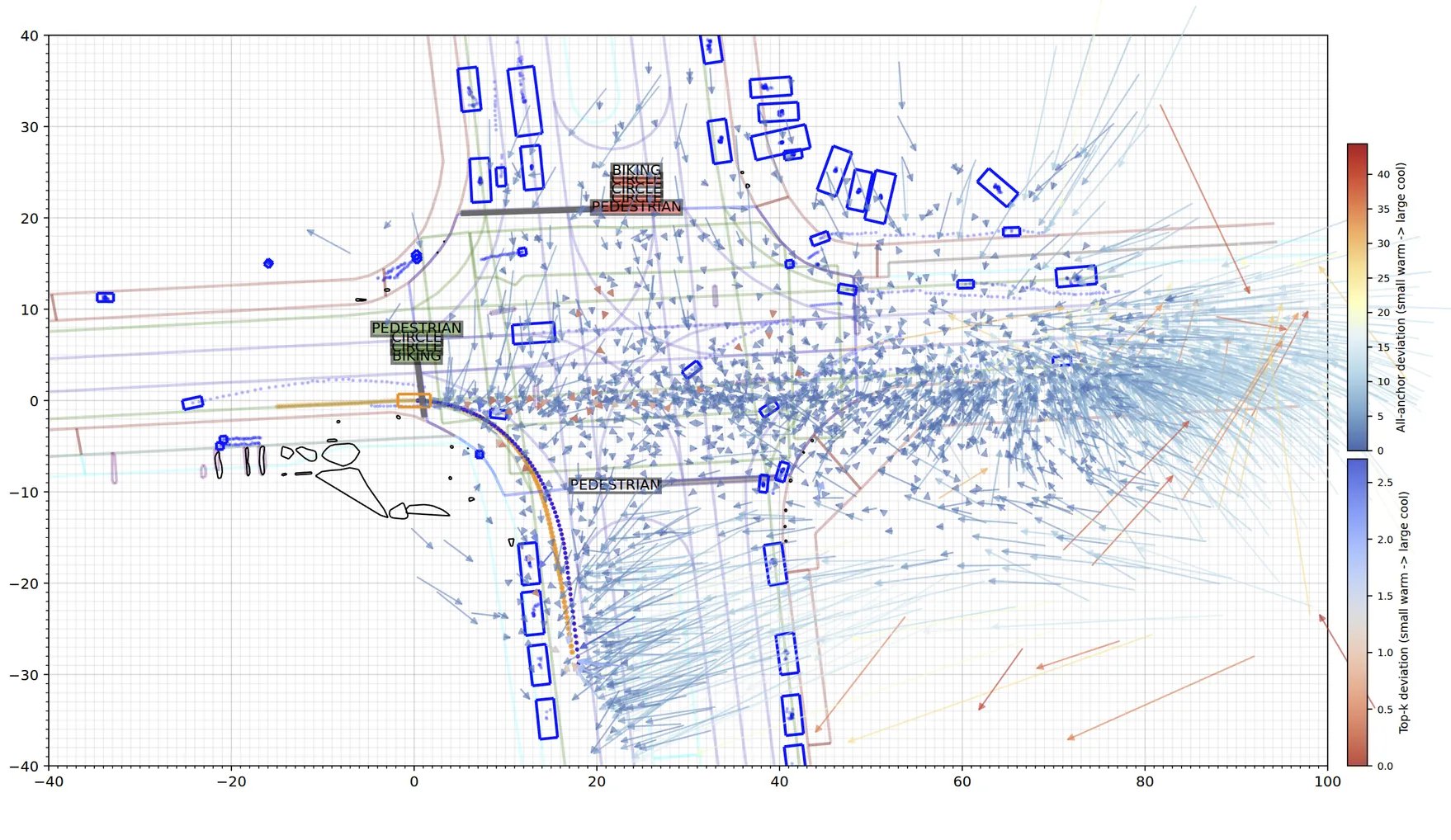}
        \caption{With local input noise}
        \label{fig:app_multimodal}
    \end{subfigure}
    \hfill
    \begin{subfigure}[t]{0.48\linewidth}
        \includegraphics[width=\linewidth, height=5.5cm, keepaspectratio]{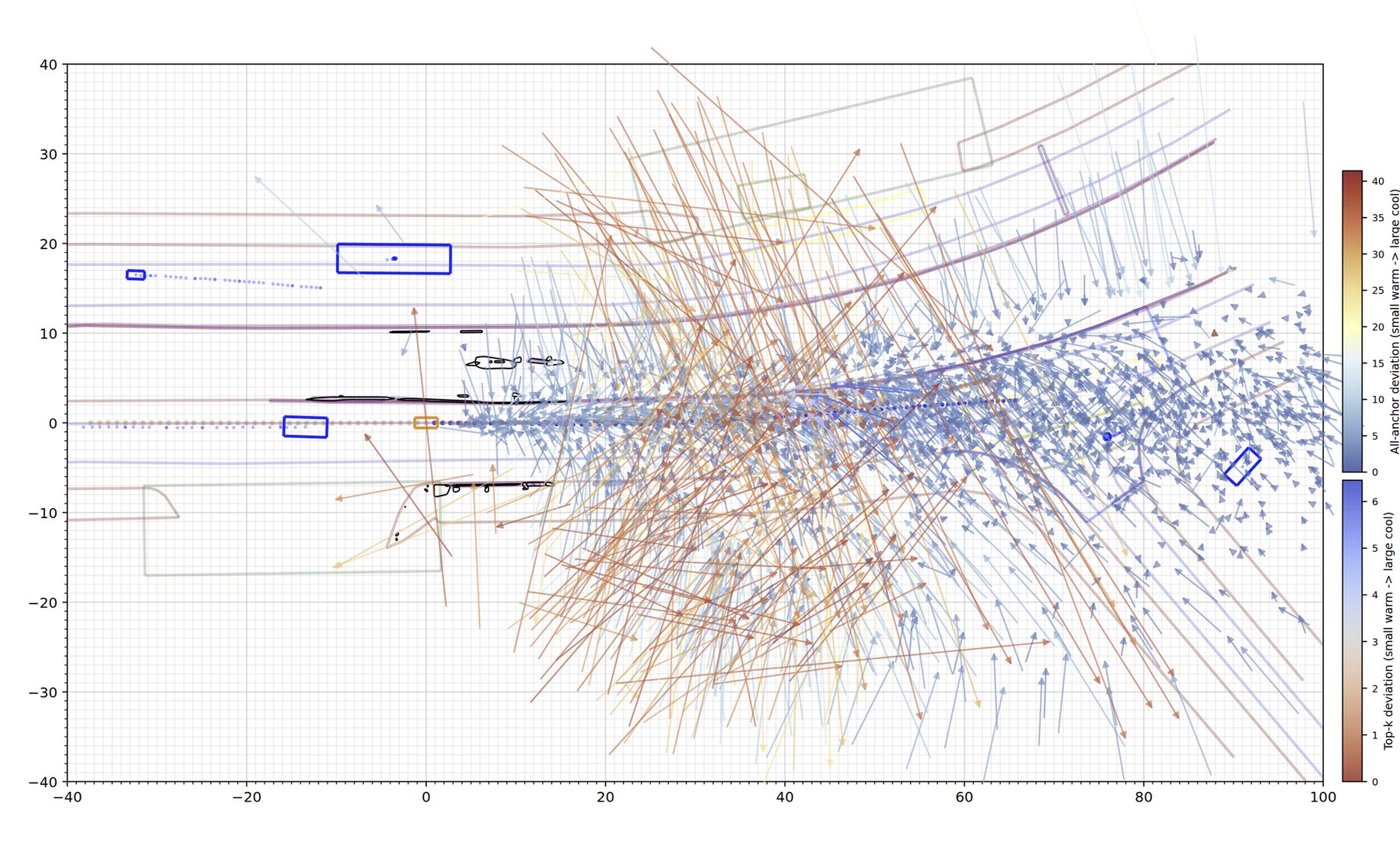}
        \caption{Without local input noise}
        \label{fig:app_knn_collapse}
    \end{subfigure}
    \caption{Stage~1 examples from a separate local-noise comparison. The left configuration uses ball perturbations and produces distinct trajectory modes; the right omits these perturbations and produces more concentrated outputs. These qualitative examples do not isolate the effect of the Voronoi acceptance rule or quantify mode coverage. Spatial coordinates are in meters.}
    \label{fig:app_stage1_viz}
\end{figure}

\paragraph{Input perturbations.}
The vocabulary defines a Voronoi partition (Figure~\ref{fig:voronoi_compare}). For each selected anchor $a$ in the $d=2T$ dimensional training representation, we use midpoint nearest-neighbor tests to construct an approximate neighbor set $\mathcal B(a)$ and set the proposal scale. Let $d_{\min}(a)$ denote the distance to the nearest other anchor in the full vocabulary, and $d_{\max}(a)$ the largest distance to a neighbor in $\mathcal B(a)$. The radii are
\[
 \begin{aligned}
 R_{\max}&=\sqrt d,\qquad R_{\rm in}=\min\{0.49d_{\min}(a),R_{\max}\},\\
 R_{\rm prop}&=\max\{R_{\rm in},\min\{0.5d_{\max}(a),R_{\max}\}\}.
 \end{aligned}
\]
Sample independent $\xi\sim\mathcal N(0,I_d)$ and $U\sim\operatorname{Uniform}(0,1)$, then propose
$y=a+R_{\rm prop}U^{1/d}\xi/\|\xi\|_2$, which is uniform within the proposal ball. Accept $y$ only if an exact Euclidean 1-NN query over the full fixed vocabulary returns $a$; otherwise retry, up to eight attempts. If every attempt fails, sample uniformly from the ball of radius $R_{\rm in}$. The approximate set $\mathcal B(a)$ therefore determines the proposal scale, while the full-vocabulary query verifies cell membership. Missing approximate neighbors cannot weaken this acceptance check. The fallback also remains in the source cell because $R_{\rm in}\le d_{\min}(a)/2$. This guarantee applies only to the perturbed start, not to all interpolated $x_t$ or to driving safety. The proposal radius approximates the cell's scale rather than its boundary, and finite rejection sampling with fallback is not uniform over the entire cell. The figure illustrates these geometric regions, not empirical sample densities; Figure~\ref{fig:app_stage1_viz} shows trajectory examples from the local-noise comparison.

\begin{figure}[htbp]
    \centering
\includegraphics[width=\linewidth, height=0.42\textheight, keepaspectratio]{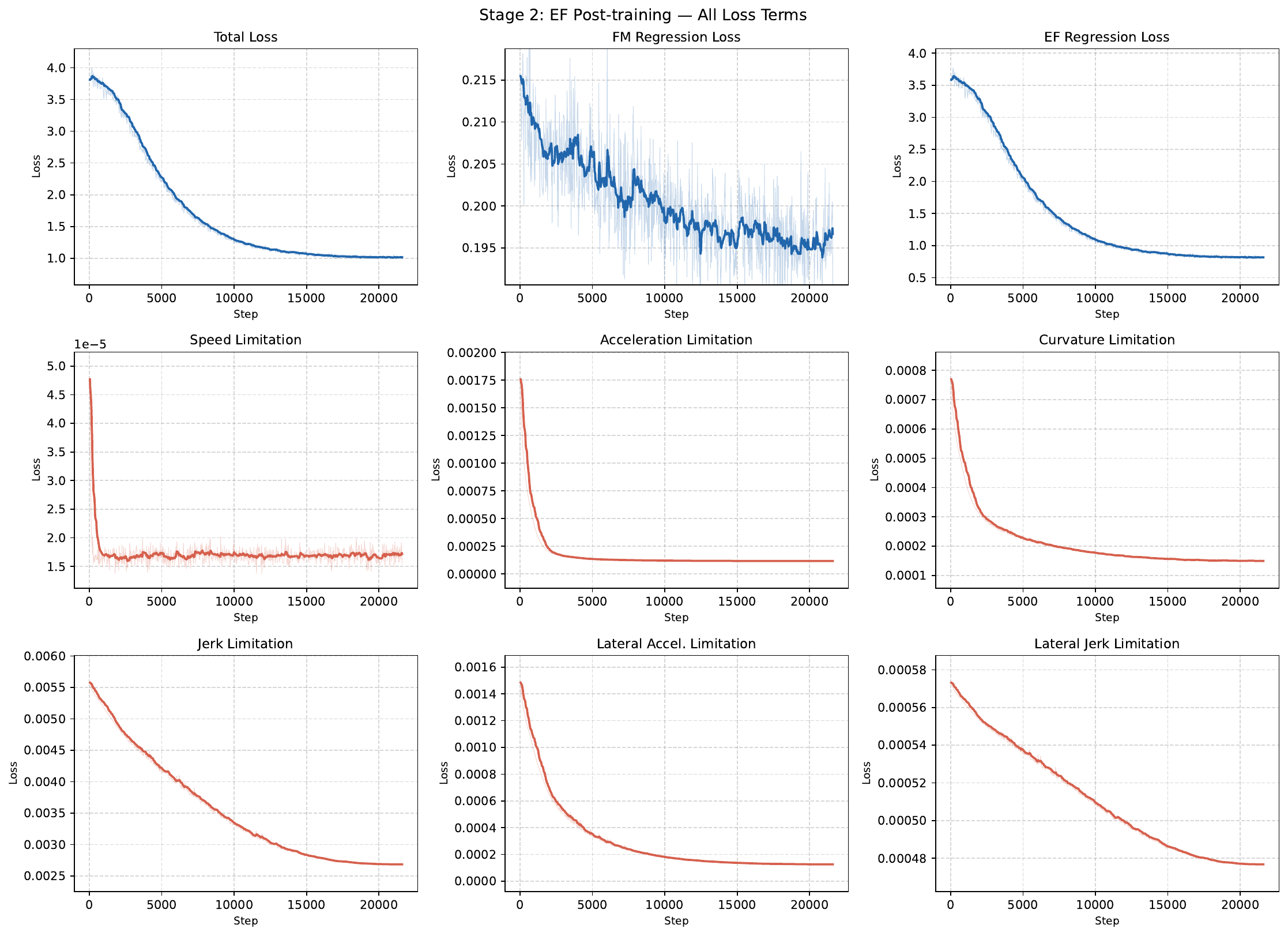}
    \caption{Stage~2 EF post-training losses over 21.6K steps using 7.5M samples.
    \textbf{Top row} (blue): total, FM regression, and EF regression losses.
    \textbf{Bottom two rows} (red): six kinematic losses for violations of speed, acceleration, curvature, jerk, lateral acceleration, and lateral jerk limits.}
    \label{fig:app_stage2_all_loss}
\end{figure}

\section{Expectation of the Anchor-Guided Estimator}
\label{app:gi_expectation}
For the current query, write $\lambda=\sigma_i$. The following analysis conditions on its retrieved neighbors, directions, and scale, treating them as fixed and excluding derivatives through their dependence on the query.
Fix a scene $s$, query $x=a_i$, and nonzero neighbor directions $u_j=u_{ij}$.
Assume that the reward is bounded and measurable.
For $h_j(q)=R(s,x+qu_j)$ and the Gaussian density
$\phi_\lambda(v)=(2\pi\lambda^2)^{-1/2}\exp(-v^2/(2\lambda^2))$, write
\[
 \mathcal Q_{j,\lambda}(\alpha)=\int_{\mathbb R}h_j(q)\phi_\lambda(q-\alpha)\,dq.
\]
Differentiating the Gaussian kernel is justified by boundedness of $h_j$ and integrability of the kernel derivative. At $\alpha=0$,
\[
 \mathcal Q'_{j,\lambda}(0)=\int h_j(q)\frac{q}{\lambda^2}\phi_\lambda(q)\,dq
 =\frac1\lambda\mathbb E_z[h_j(\lambda z)z]
 =\frac1\lambda\mathbb E_z[(h_j(\lambda z)-h_j(0))z].
\]
This proves Eq.~\eqref{eq:gi_expectation} without differentiating the collision oracle.
The factor $z$ arises from differentiating the Gaussian kernel and weights the reward difference by the signed radial sample.

\paragraph{Meaning and ascent of the local surrogate.}
For this appendix derivation, we rename the query $a_i$ as $x$ and the retrieved directions $u_{ij}$ as $u_j$. Eq.~\eqref{eq:local_surrogate} can then be written as
$H_x(\Delta)=N^{-1}\sum_j\mathcal Q_{j,\lambda}(u_j^\top\Delta)$.
The chain rule and the kernel identity above give
\[
 \nabla_\Delta H_x(0)=\frac1N\sum_j\mathcal Q'_{j,\lambda}(0)u_j
 =\mathbb E[\hat g\mid s,x,\{u_j\}].
\]
Convolving a bounded reward with a one-dimensional Gaussian gives a smooth function. With the query and direction pool fixed, $H_x$ is therefore smooth in $\Delta$. Let $h=\nabla H_x(0)$. Then
\[
 H_x(\eta h)=H_x(0)+\eta\|h\|^2+O(\eta^2).
\]
Thus, following the expected direction in trajectory space locally increases $H_x$ whenever $h\ne0$ and $\eta>0$ is sufficiently small. Individual noisy updates need not improve it monotonically.

\paragraph{Connection to parameter updates.}
Fix a training batch $\mathcal B$ and reference parameters $\theta_0$. Write $a_i(\theta)=T_\theta(x_{0,i},s_i)$ and $x_i=a_i(\theta_0)$, and hold each query's retrieved neighbors, directions, and scale fixed. The corresponding parameter-space surrogate for this batch is
\[
 \mathcal H_{\theta_0,\mathcal B}(\theta)
 =\frac1{|\mathcal B|}\sum_{i\in\mathcal B}
 H_{x_i}\bigl(a_i(\theta)-x_i\bigr).
\]
At a differentiability point of the decoder, the chain rule gives the exact conditional identity
\[
 -\mathbb E_z[\nabla_\theta\mathcal L_{\rm RL}(\theta_0)]
 =\frac1{|\mathcal B|}\sum_i
 D_\theta a_i(\theta_0)^\top\nabla H_{x_i}(0)
 =\nabla_\theta\mathcal H_{\theta_0,\mathcal B}(\theta_0).
\]
The stop-gradient implementation therefore estimates the gradient of this fixed batch surrogate, accounting for the shared parameters. As the reference parameters change, the surrogate also changes; the identity does not establish one global objective for the entire training process. Let $g=\nabla\mathcal H_{\theta_0,\mathcal B}(\theta_0)$ and $\ell=\nabla\mathcal L_{\rm FM}(\theta_0)$, where the latter includes the retained kinematic term. For an ordinary gradient step, the expected combined direction is $d=\beta g-\ell$, and the first-order change in the surrogate is
\[
 g^\top d=\beta\|g\|^2-g^\top\ell.
\]
A strict first-order gain requires positive alignment between the surrogate gradient and the combined update direction. A finite-step improvement bound in expectation additionally requires smoothness, a sufficiently small step, and controlled estimator variance.

Define the reward-survival event $C_t(y)=\{R(s,y)>t\}$ for $t=0,\ldots,T$. For $t<T$, it means no collision through predicted-waypoint index $t$. The sentinel convention gives $C_T=C_{T-1}$, both denoting safety over all $T$ waypoints; $T$ is not an additional checked waypoint. The integer collision-step reward in Eq.~\eqref{eq:reward} admits the tail-sum representation
\[
 R(s,y)=\sum_{t=0}^T\mathbf1\{C_t(y)\},\qquad
 H_x(\Delta)=\frac1N\sum_{j=1}^N\sum_{t=0}^T
 \Pr_z\{C_t(x+(u_j^\top\Delta+\lambda z)u_j)\}.
\]
Thus, $H_x$ sums collision-free probabilities over the checked waypoints, with one additional full-horizon term from the sentinel convention, so it also rewards delaying a collision. The full-horizon collision-free output fraction under this oracle is
$J_{\rm safe}(\theta)=\mathbb E_{s,x_0}\mathbf1\{C_T(T_\theta(x_0,s))\}$.
Increasing the collision-step surrogate does not necessarily increase the probability of remaining collision-free over the full horizon. If the reward were instead $S(s,y)=\mathbf1\{C_T(y)\}$, the same estimator identity would hold, and the local surrogate would equal the average collision-free candidate probability. This alternative illustrates the identity; it is not the reward used in our implementation.

\paragraph{Conditional variance.}
If $|R(s,y)-R(s,x)|\le B$ for all queried $y$, independent radial draws give
\[
 \mathbb E\|\hat g_i-\mathbb E\hat g_i\|_2^2\le\frac{B^2}{N\lambda^2}.
\]
Here all expectations are conditional on the fixed query and direction pool. For the reward in Eq.~\eqref{eq:reward}, the conservative bound $B=T+1$ also covers collisions at step zero.
For a composite reward with additional kinematic penalties, the expectation identity applies under the stated boundedness assumption, but the collision-probability interpretation and the choice $B=T+1$ do not transfer. A bound for the actual composite reward must be established separately.
Using $m$ independent draws per direction changes the denominator to $Nm\lambda^2$ and requires $Nm+1$ oracle queries.

\paragraph{Smooth-reward special case.}
If $R(s,\cdot)$ has a globally bounded third derivative with operator norm at most $L_3$, Taylor expansion gives
\[
 \mathbb E\hat g_i=G_i\nabla R(s,a_i)+b_i,\qquad
 G_i=\frac1N\sum_j u_{ij}u_{ij}^\top,\qquad
 \|b_i\|_2\le\tfrac12L_3\lambda^2.
\]
The factor $1/2$ follows from $\mathbb E[z^4]/6=1/2$.
$G_i$ is positive semidefinite with trace one and rank at most $N$; it is generally not an orthogonal projector. The rank bound follows because its range lies in $\operatorname{span}\{u_{ij}\}_{j=1}^N$. This is a restriction on the query-local output-space signal, not on all effects of a shared-parameter network update. This Taylor statement does not apply to the raw collision-step reward.

\paragraph{Scope of the result.}
Eq.~\eqref{eq:local_surrogate} defines a scalar objective for a fixed query. Updating the query and retrieving new neighbors changes this local objective, and the resulting objectives need not correspond to one globally smoothed function. Moreover, probing each direction separately need not capture the interactions in the combined trajectory $x+\Delta$. Ascent of the local surrogate alone therefore guarantees neither an increase in the true expected reward nor a larger safe-output fraction. Global ZO-SGD convergence would require a separate argument.

\paragraph{Exploration subspace.}
For a fixed query, the estimate lies in the span of the retrieved directions, whose dimension is at most 16 when $N=16$ in the 160-dimensional trajectory space. Orthogonal directions receive no direct signal from this query. The following feasibility analysis concerns speed and acceleration constraints; it does not establish obstacle avoidance or feasibility for every model-generated query.

\paragraph{Directional feasibility probability.}
Convex combinations of trajectories satisfying the same convex finite-difference constraints are feasible, but signed Gaussian extrapolations need not remain in their convex hull. More generally, for any query $x$ and constraint set $\mathcal F$, a certified interval
$\{x+qu_j:L_j\le q\le U_j\}\subseteq\mathcal F$ gives
\begin{equation}
 \Pr\{x+\lambda zu_j\in\mathcal F\}
 \ge\Phi(U_j/\lambda)-\Phi(L_j/\lambda).
 \label{eq:direction_feasibility}
\end{equation}
Here $\Phi$ is the standard normal CDF. Equality holds when $[L_j,U_j]$ is the entire feasible intersection with the line; an empty intersection has probability zero. When $L_j\le0\le U_j$, write $L_j=-r_j^-$ and $U_j=r_j^+$ for nonnegative radii to obtain a violation bound
$\Phi(-r_j^-/\lambda)+\Phi(-r_j^+/\lambda)$.
The bound follows because $L_j\le\lambda z\le U_j$ implies feasibility. Averaging over directions and queries bounds the expected fraction of feasible candidates, but individual batches may differ from that expectation.

For position trajectories with uniform step $h$, internal speed and acceleration constraints take the form
\[
 \|D_1y\|_{2,\infty}\le25,\qquad
 \|D_2y\|_{2,\infty}\le4.5,
\]
where $D_1$ and $D_2$ include $h^{-1}$ and $h^{-2}$, respectively. Each time-indexed constraint along $y=x+qu_j$ is a scalar quadratic inequality
$\|b_t+qc_t\|^2\le c_{\max}^2$.
Intersecting its feasible intervals over time and constraint type gives $[L_j,U_j]$ exactly, or an empty set. No isotropy or Gaussian assumption about the anchor directions is needed; the Gaussian is the radial sampling construction. The argument also applies to third-difference jerk constraints if those definitions are adopted, but jerk is not evaluated here.

\end{document}